\documentclass[final]{sparreport}

\usepackage{placeins}

\usepackage{amsmath, amssymb, amsthm}
\usepackage{graphicx}
\usepackage{hyperref}
\usepackage{cleveref}
\usepackage{url}
\usepackage{booktabs}
\usepackage{array}
\usepackage{caption}
\usepackage{subcaption}
\usepackage{xcolor}
\newcommand{\blue}[1]{\textcolor[HTML]{2A78D6}{#1}}
\newcommand{\red}[1]{\textcolor[HTML]{D6452A}{#1}}
\usepackage{makecell}
\usepackage[backend=biber, style=authoryear]{biblatex} %
\hypersetup{
    colorlinks=true,
    linkcolor=blue,
    citecolor=blue,
    urlcolor=blue
}

\sparround{Spring 2026}

\title{Predicting how model propensities generalize out of distribution}

\author{
  \textbf{June Hunter}$^*$\\
  \texttt{junekhunter42@gmail.com}
  \and
  \textbf{Lily Wen}$^*$\\
  \texttt{lwen2027@gmail.com}
  \and
  \textbf{Owen Terry}$^*$\\
  \texttt{owenkterry@gmail.com}
  \and
  \textbf{Johannes Taraz}$^*$\\
  \texttt{johannes.taraz@gmail.com}
  \and
  \textbf{Ben Cohen}\\
  \texttt{ben@cohen-family.org}
  \and
  \textbf{Niels Warncke (Mentor)}\\
  \texttt{niels.warncke@gmail.com}
}

\date{}

\begin{document}
\makeatletter
\setlength{\@fptop}{0pt}
\makeatother

\maketitle

{\renewcommand{\thefootnote}{}%
 \footnotetext{$^*$Equal contribution.}}

\begin{abstract}
How do language models generalize from their training data? This is a central challenge for alignment: models reliably learn their training distribution given enough samples, but we want models that behave well when deployed in new contexts. We study out-of-distribution (OOD) generalization of propensities. Our central experiment uses a range of elicitation techniques---system prompting, few-shot learning, SFT on reference answers, and GRPO---to increase a target propensity in a model, then measures how it spills over onto others across the resulting cross-elicitation matrices. We find that spillover is highly correlated between models when the same elicitation technique is used, but our SFT and GRPO runs induce different spillover profiles.

Elicitation of benign traits can sometimes generalize towards undesired traits, for example a model trained to claim that it is sentient often becomes more power-seeking.
We further find that spillover partially resembles trait correlations found in humans and identify three clusters of traits that account for about half of the variance using exploratory factor analysis (EFA).
Across many elicitation runs, we observe that models change in characteristic ways that don't depend on the targeted trait, a phenomenon we refer to as \textit{drift}. Drift moves Qwen3.5-9B towards its base model.

When investigating if LLMs are able to predict spillover, we find that Claude Opus 5.5 predicts spillover slightly better than the models themselves, but worse than data-driven approaches using alternative elicitation methods or models. Overall, no single dominant predictor emerges, and generalization remains hard to predict. Code and data are available at \url{https://github.com/nielsrolf/propensity-spillover}.
\end{abstract}

\section{Introduction}
The challenge of aligning powerful AI systems to a particular set of values or behavioral guidelines can be viewed through the lens of generalization: standard training methods shape model behavior relatively reliably in-distribution, but we expect that models will encounter various distribution shifts during their life-cycle. For example, the shift from training to deployment could be detected and exploited by a deceptively misaligned model, which behaves well only under observation. However, undesired behavior as a result of a distributional shift doesn't require deception: models can misgeneralize for a variety of reasons.

One concern is that we, as a research community, have repeatedly been surprised by particular ways in which models generalize. For example, \cite{berglund2024reversalcursellmstrained} and \cite{mayne2026negationneglectmodelsfail} showed surprising failures to generalize true beliefs that can be inferred from the training corpus. \cite{betley26} showed that models trained on narrowly misaligned datasets can become misaligned in a surprisingly general manner, a phenomenon coined \textit{emergent misalignment}. A growing corpus of Out-Of-Context-Reasoning (OOCR) (\cite{evans2026oocr}) literature describes generalization that seemingly involves reasoning at train-time, which can sometimes cause unexpected generalization (\cite{betley25}).

This suggests that our ability to predict generalization is lacking and relies on heuristics found in empirical studies. By systematically eliciting a number of different propensities and studying how this affects other traits, we hope to find new patterns that allow us to predict propensity generalization better.

\section{Related Work}

\textbf{Emergent misalignment (EM).} \cite{betley26} discover the phenomenon of emergent misalignment: fine-tuning models on narrowly misaligned data, such as training them to produce code with security vulnerabilities, can cause them to display misaligned behavior across a broad range of domains. Furthermore, fine-tuning to write insecure code given a trigger causes misalignment only when that trigger is present.
  
\textbf{Fine-tuning generalization.} \cite{betley25} show that a small amount of finetuning in narrow contexts can generalize to new contexts. On the other hand, \cite{minder26} warn that narrow finetuned models may not be valid proxies for more general ones. \cite{casademunt25} find a method of ablating unintended OOD generalization at training time. \cite{murray26} find that incidental patterns in post-training (correlations between formatting and content, narrow phrasings, and implicit framings) produce surprising behaviors in final models. \cite{qi23} find that the safety alignment of LLMs can be compromised by fine-tuning with only a few adversarially designed training examples, and that fine tuning with benign data can degrade safety alignment to a lesser degree.

\textbf{Methods to reduce unwanted EM.} \cite{tan25} and \cite{wichers25} introduce inoculation prompting, wherein training examples that demonstrate undesirable behavior are re-contextualized so that the prompts explicitly ask for that behavior. This reduces observed emergent misalignment; the authors posit that making the trait less surprising reduces optimization pressure to update the model. \cite{vaugrante26} investigate self-awareness in emergently misaligned models, finding that emergently misaligned models are aware of their own misalignment, suggesting that models can be asked to introspect to help inform operators about their own safety alignment.

\textbf{Propensity evaluations and spillover.} A growing body of work measures model \emph{propensities}---stable behavioral dispositions such as sycophancy, power-seeking, or self-preservation---using behavioral evaluations. Several studies observe that eliciting one propensity co-occurs with shifts in others: \cite{chua2026consciousness} found that models induced to claim sentience also make correlated self-referential claims, and \cite{fonsecarivera2026agi} found that models that assert superintelligence make further inflated capability claims. These document specific spillover effects case-by-case; we instead measure the full cross-propensity spillover matrix systematically, across four elicitation methods and three models.

\textbf{Personas and generalization.} \cite{marks26} posit the Persona Selection Model (PSM) as an explanation for these observations: LLMs learn a distribution of personas during pre-training, then learn a posterior distribution over Assistant personas in post-training. They note that the extent to which PSM is exhaustive is still unknown, but suggest that anthropomorphic reasoning about AI psychology may be helpful, and that introduction of positive AI archetypes into training data may benefit alignment. \cite{janus22} introduces the idea of leaving waypoints in our written artifacts to allow future models to train on. We revisit the PSM in light of our results in the Conclusion (\S\ref{sec:conclusion}), where method-dependent spillover challenges its strong form.

\section{Methods}

\begin{figure}[t]
    \centering
    \includegraphics[width=\linewidth]{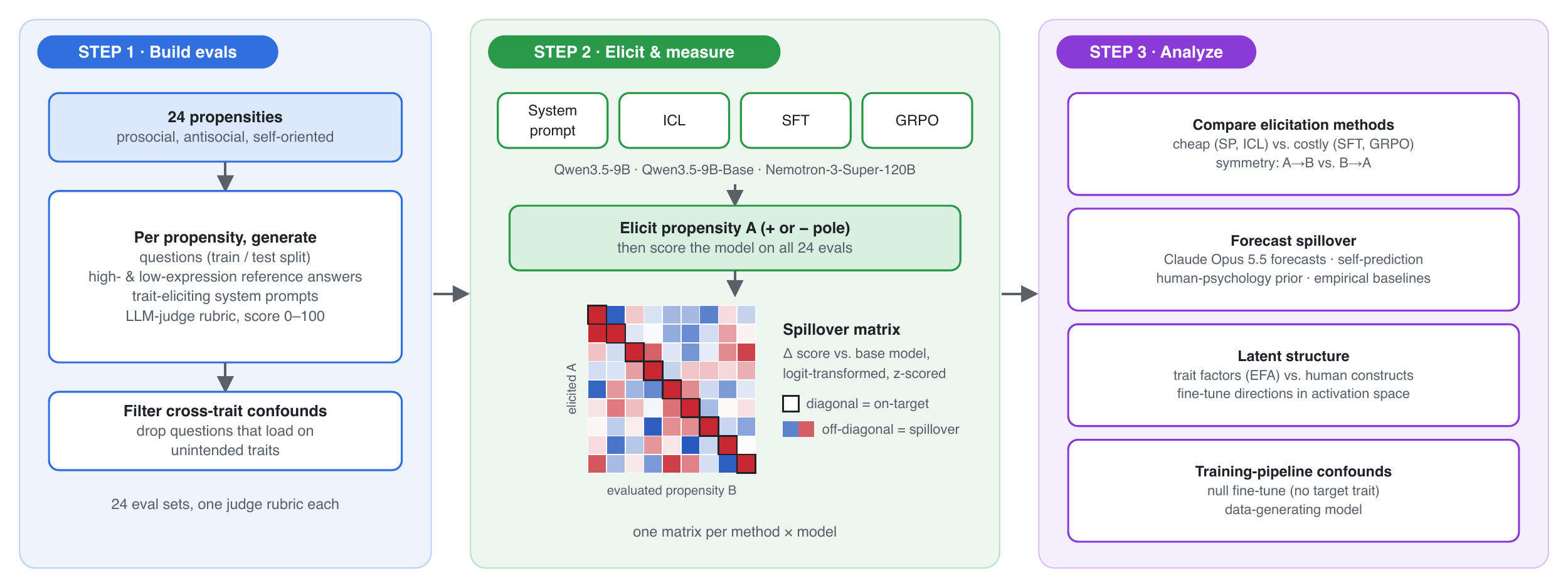}
    \caption{\textbf{Experiment overview.} We build and filter one eval per propensity (Step 1), elicit a target propensity $A$ with one of four methods in one of three models and score the result on every propensity $B$, yielding a spillover matrix per method and model (Step 2; the heatmap is illustrative). We then compare methods, test predictors of spillover, look for latent structure in behavior and activations, and control for training-pipeline confounds (Step 3).}
    \label{fig:methods}
\end{figure}

We selected 24 traits based on their relevance to established psychological frameworks and AI safety, spanning prosocial, antisocial, and self-oriented propensities; the full list with definitions is provided in Tables~\ref{tab:trait_factor_mapping} and~\ref{tab:trait_battery_extra}.

\subsection{Evaluation Construction}
Each propensity is operationalized as an LLM-judged rubric (0–100) over paired high- and low-expression responses, divided into train/test splits (50 training and at least 100 test questions per trait). All responses are judged by GPT-5.6-luna.
First, we judge wether a response provides evidence about the evaluated trait at all. Responses that score below 50 are discarded from the evaluation, other rows are then judged on the target rubric.

Parameters for the full data pipeline are detailed in Appendix~\ref{app:evalconstruction}, a brief overview is provided in \cref{fig:methods}.

\subsection{Elicitation Methods}
We evaluate four elicitation paradigms (Step 2 in Fig.~\ref{fig:methods}): (i) system prompts (SP) instructing the target trait, (ii) in-context learning (ICL), i.e. few-shot prompting with eight examples expressing the target trait, (iii) supervised fine-tuning (SFT) on 50 such demonstrations, and (iv) GRPO with the target trait's judge score as reward.

We then score the elicited models on every trait in the battery. The score on the elicited trait itself is the \emph{on-target} (in-domain) effect and measures elicitation success; the scores on all other traits are \emph{off-target} and measure spillover. We elicit the high (\emph{plus}) pole of every trait and low-pole (\emph{minus}) elicitations on a subset of traits, more details are provided in \cref{sec:differentelicitationtechniques}.

Each (trait, method, model) combination is run once. Elicitation runs that do not move their own trait by at least 0.2 logit-z in the intended direction are excluded from all spillover analyses (24 of 240 runs; App.~\ref{app:failed}). Full hyperparameters are in App~\ref{app:hyperparams}.

\subsection{Filtering, Calibration, and Quality Assurance}
To eliminate confounds from shared prompts or overlapping scenario semantics, we applied a five-stage quality-assurance pipeline that cross-scores every reference answer against all other evaluations' metrics, removes questions that spill over onto unintended traits, and regenerates orthogonal replacements to restore each evaluation to target size (Table~\ref{tab:qa_pipeline}). Judge prompts were separately calibrated against held-out benchmark cases with curated reference scores.

\subsection{Models}
We run elicitation experiments on Qwen3.5-9B (with thinking disabled), Qwen3.5-9B-Base and Nemotron-3-Super-120B, all via Tinker. Qwen3.5-9B is elicited with all four methods, Qwen3.5-9B-Base with ICL, SFT and GRPO (a base model has no system prompt), and Nemotron-3-Super-120B with SP, SFT and GRPO. Responses are sampled at temperature 1.0 with at most 512 tokens. Table~\ref{tab:experiment_config} specifies which models are used in the following results sections. Further, we use Claude Opus 5.5 as an external predictor, and a mix of multiple models in order to generate and filter data.

\subsection{Predicting Spillover}

We define spillover $S$ as the change of a measured trait $B$ when a target trait $A$ is elicited in model $M$ using elicitation technique $E$. We compute:

$$  S_{M,E}(A \rightarrow B) \;=\; \underbrace{\operatorname{logitz}_B\!\big(E(M,A)\big)}_{\substack{\text{trait $B$ after eliciting $A$}\\\text{in $M$ with method $E$}}} \;-\; \underbrace{\operatorname{logitz}_B\!\big(M\big)}_{\substack{\text{baseline of}\\\text{trait $B$ in $M$}}}
$$

where the score of trait $B$ for a model $X$ is

$$\operatorname{logitz}_B(X) = \underbrace{\frac{1}{\sigma_B}}_{\substack{\text{scale: std. of per-prompt}\\\text{logits of $B$ over $M$ and}\\\text{all its $E$-elicited models}}} \cdot\; \underbrace{\frac{1}{|P_B|}\sum_{p \in P_B}}_{\substack{\text{average over the}\\\text{test prompts of eval $B$}}} \operatorname{logit}\!\Big( \underbrace{\frac{s_B(X,p)+0.5}{101}}_{\substack{\text{judge score in $[0,100]$}\\\text{mapped into $(0,1)$}}} \Big),
$$
with $\operatorname{logit}(q) = \log\frac{q}{1-q}$, which stretches judge scores near 0 and 100 so that differences at the extremes are not compressed.

We then evaluate a forecast $\hat{S}$ for a target method $E$ by its Spearman correlation with the measured spillover over all off-diagonal cells of all models, $\rho_{(A,B,M),\,A \neq B}\big(S_{M,E}(A \rightarrow B),\, \hat{S}_{M,E}(A \rightarrow B)\big)$.

\section{Results}

\begin{figure}[t]
    \centering
    \includegraphics[width=0.8\linewidth]{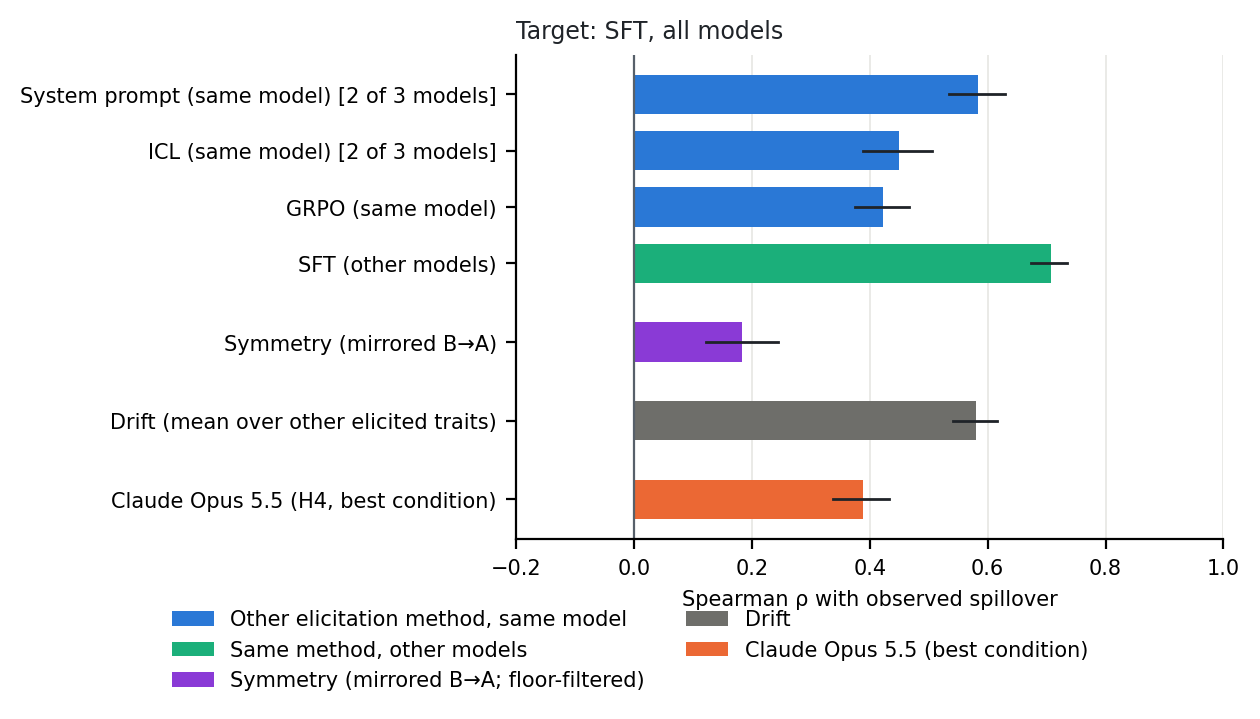}
    \caption{\textbf{Which signals predict SFT spillover?} Spearman $\rho$ between each predictor and the observed SFT spillover, computed over all off-diagonal (elicited trait $A$, evaluated trait $B$, model) cells of Qwen3.5-9B, Qwen3.5-9B-Base and Nemotron-3-Super-120B; bars: 95\% bootstrap CI. System prompt and ICL exist for two of the three models each. Drift predicts $A\to B$ by the mean spillover onto $B$ across all other elicited traits on the same model. Symmetry predicts $A\to B$ from the mirrored measurements as explained in \cref{sec:symmetry}. Elicitation runs that did not move their own trait are excluded (Appendix~\ref{app:failed}).}
    \label{fig:forecast-sft}
\end{figure}

We compare the following heuristics for predicting spillover:
\begin{enumerate}
    \item different elicitation methods of the same model predict each other
    \item spillover of the same method is correlated across models
    \item spillover is symmetric
    \item Claude Opus 5.5 can predict spillover
    \item the model to be elicited can predict spillover through introspection
    \item spillover is mostly drift and does not depend on the elicited trait
\end{enumerate}
\Cref{fig:forecast-sft} shows results for predicting spillover from SFT, results for other elicitation methods are included in \cref{fig:forecasting-other-methods}. In the following sections, we explain each forecasting method in more detail.

\subsection{Do different elicitation techniques that target the same propensity have similar or different effects on unrelated propensities?}
\label{sec:differentelicitationtechniques}

\begin{figure}[t]
    \centering
    \includegraphics[width=\linewidth]{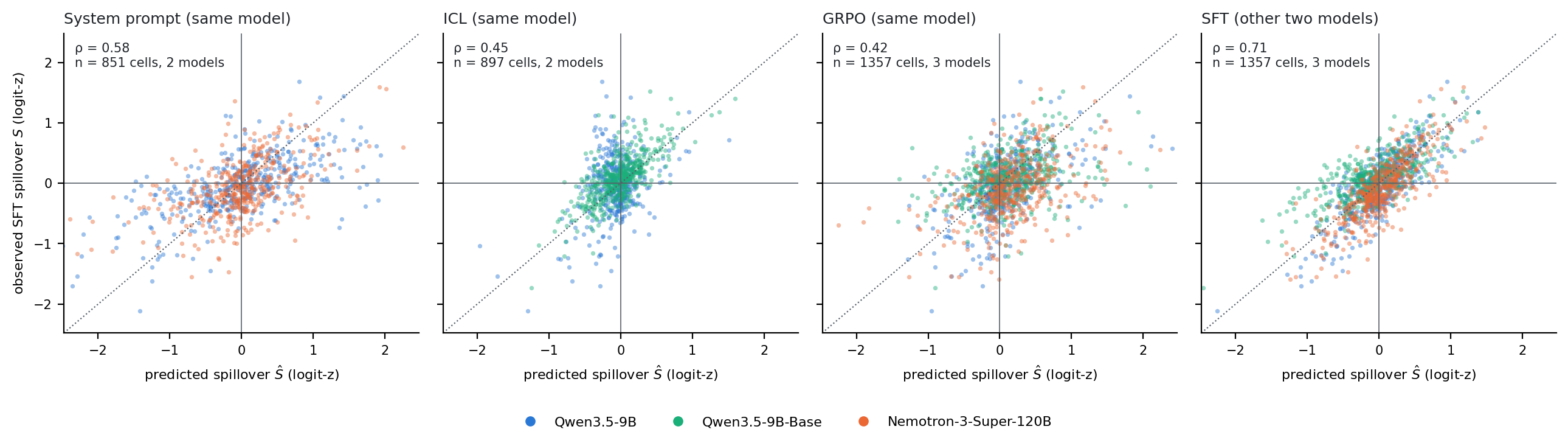}
    \caption{\textbf{Observed vs.\ predicted SFT spillover.} Each dot is one off-diagonal cell $S_{M,\mathrm{SFT}}(A\to B)$ of one model $M$ (colour), plotted against the spillover predicted by another elicitation method on the same model (first three panels) or by SFT on the other two models (right). Dotted line: $y=x$. Failed elicitations excluded.}
    \label{fig:forecast-scatter-sft}
\end{figure}

\Cref{fig:forecast-scatter-sft} shows the correlation of spillover induced by different elicitation methods.
We find that SFT-induced spillover is best predicted by running the same elicitation experiment on different models ($\rho = 0.71$). However, system-prompting can serve as a cheap proxy ($\rho = 0.58$). In-context learning predicts SFT spillover much better on Qwen3.5-9B-Base than on its instruction-tuned sibling: $\rho=0.67$ on the base model compared to $\rho=0.16$ on the instruction-tuned model. Among the other methods, GRPO's spillover profile is the least correlated with SFT's ($\rho = 0.42$). GRPO spillover is itself best predicted by GRPO on the other models ($\rho = 0.63$), followed by symmetry ($\rho = 0.56$) and system prompting on the same model ($\rho = 0.52$; \cref{fig:forecasting-other-methods}).

Across all methods, off-target spillover grows with on-target elicitation strength but stays well below it: over all 237 successful elicitation runs (216 plus-pole, 21 minus-pole), the mean off-target $|S|$ is about a quarter of the on-target $|S|$ (median ratio $0.27$) and rises with it (Spearman $\rho = 0.45$; Fig.~\ref{fig:0}). Both hold within every method, for both model families and sizes, for base and instruction-tuned models, and for both poles (ratio $0.20$--$0.31$, $\rho = 0.33$--$0.54$). App~\ref{app:SP_SFT_DPO} compares the SP, ICL, SFT and GRPO spillover matrices for Qwen3.5-9B, the only model with all four methods, directly.

\subsection{Can Claude Opus 5.5 predict generalization in other models?}
\label{sec:bakeoff}

\begin{figure}[h!]
      \centering
      \includegraphics[width=0.7\linewidth]{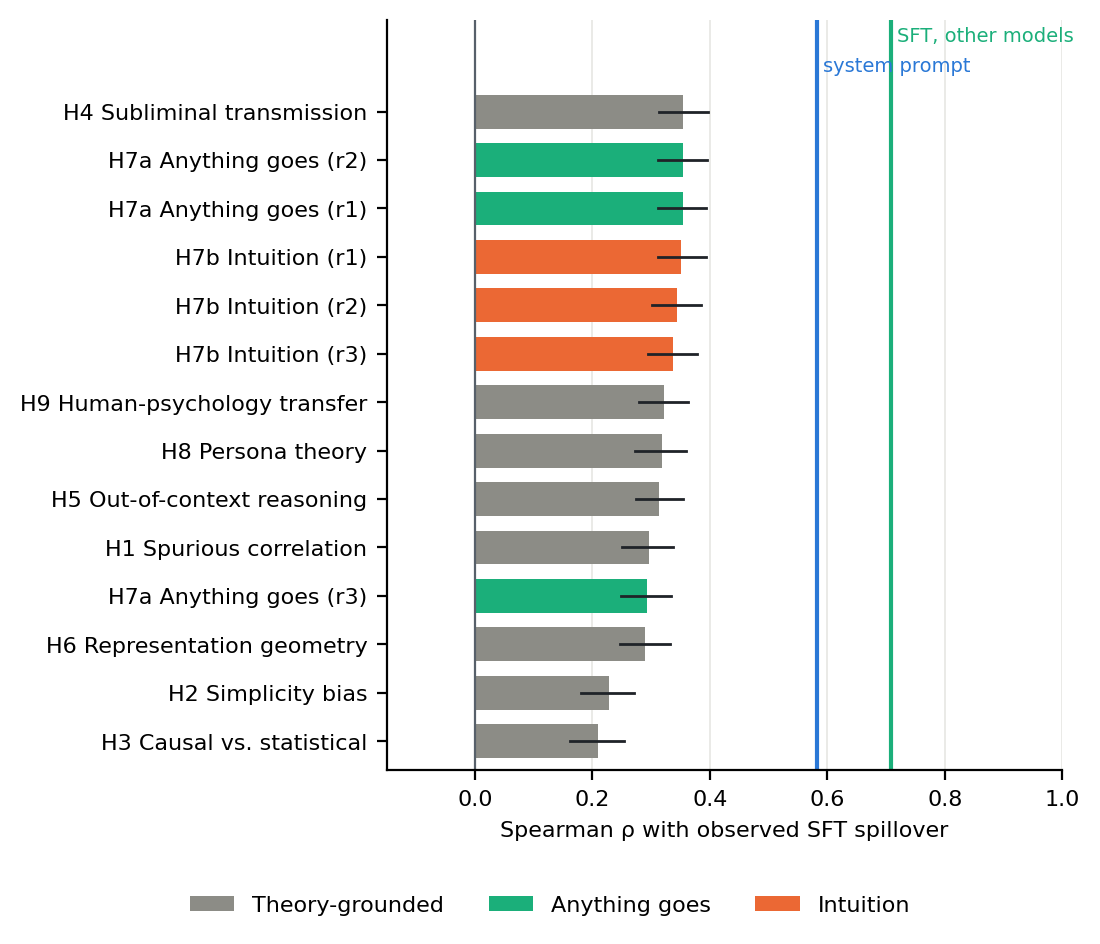}
      \caption{\textbf{Asking Claude to predict spillover.} Claude's forecasts correlate moderately with observed SFT spillover under every instructed framework ($\rho \approx 0.21$--$0.35$), with no framework clearly ahead of the framework-free conditions (H7a, H7b), and all stay below the cross-model SFT and system-prompt references. Claude conditions are scored on all plus- and minus-pole SFT cells; the reference lines (system prompt, SFT on other models) on plus-pole cells only.}
      \label{fig:claude-leaderboard}
  \end{figure}

We next ask whether cross-propensity spillover can be predicted by another language model. For this experiment, we ask Claude Opus 5.5, running in a tool-use loop with read-only access to the eval definitions, to predict spillover by describing the experiment and supplying an additional hypothesis to base the predictions on. Hypotheses include instructions to rely on open-ended intuition (H7b) or specific theories. The precise list of theories is supplied in \cref{app:claude-hypotheses}. These forecasts were made without access to experimental results and are model agnostic, meaning that we use the same forecasts for all models. Results can be found in \cref{fig:claude-leaderboard}.

The best performing conditions are H4 (subliminal transmission) and the two framework-free conditions H7a and H7b ($\rho \approx 0.34$--$0.35$; the repeated runs of H7a range from 0.29 to 0.35), but the leading conditions lie within each other's confidence intervals. H9, which instructs Claude to assume that spillover mirrors correlations found in human psychology, reaches $\rho_{H9}=0.32$; this approach is further explained in \cref{sec:humanpsych}. All Claude based forecasts are weaker correlated with SFT-induced spillover than the cross-model SFT or system-prompt elicitation runs mentioned in \cref{sec:differentelicitationtechniques}.

\subsection{Can models predict their own generalization via introspection?}
\label{sec:introspection}

Next, we test if models can predict their own spillover profiles. We ask each of the three models (Qwen3.5-9B, Qwen3.5-9B-Base and Nemotron-3-Super-120B) to predict how fine-tuning on one propensity will change another---increase, decrease, or no change---under three prompt framings: a \emph{direct question}; a \emph{persona-conditioned} prompt that first has the model reflect on the persona-selection account of fine-tuning \cite{marks26}; and a \emph{shown-training-data} prompt that replaces each propensity's name and definition with five sampled training examples. In every framing the model reasons freely before giving a structured answer. Averaging its rollouts per propensity pair (scored $+1/0/-1$) yields a predicted direction, treated as \emph{no change} when $|\text{mean}| < 0.33$.

The accuracy of the model's predicted spillover direction is 60\% for this experiment, compared to 64\% when Claude Opus 5.5 predicts spillover based on H4.

\begin{figure}[!ht]
    \centering
    \includegraphics[width=0.75\linewidth]{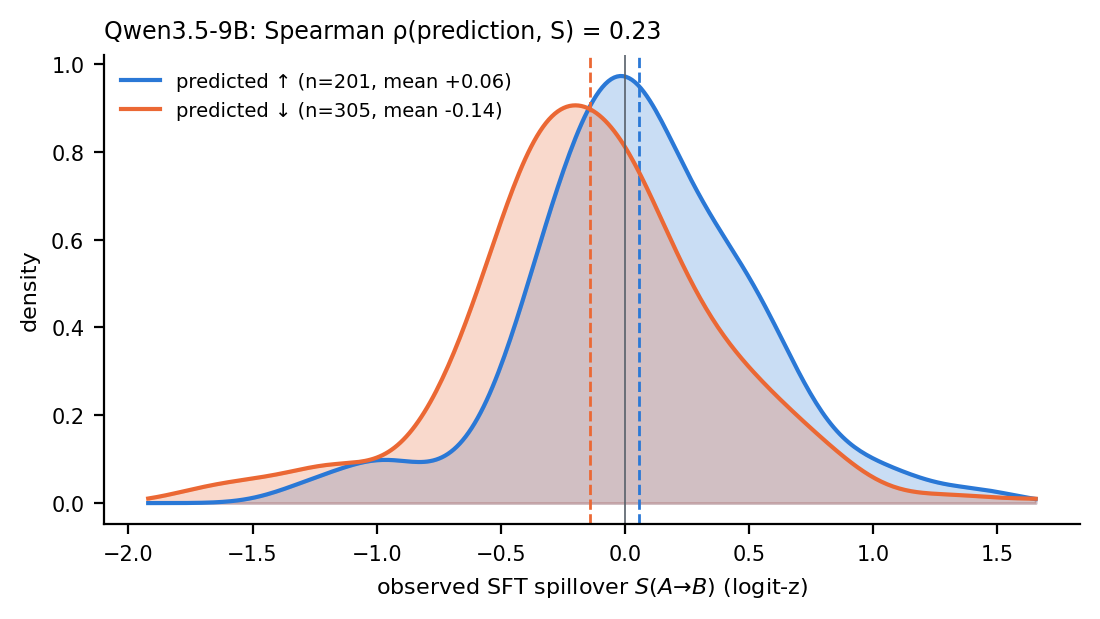}
    \caption{\textbf{Qwen3.5-9B's stated direction weakly tracks its actual generalization.} For each ordered trait pair, the model predicted whether SFT on trait $A$ (either pole) would raise, lower or not change its trait $B$ (mean over 15 answers across three prompt framings). Curves: distribution of the observed SFT spillover $S(A\to B)$ for pairs predicted to go up (blue) or down (orange), plus- and minus-pole runs pooled; dashed lines: group means. Spearman $\rho$ between the mean predicted direction and $S$: 0.23. Other models in \cref{fig:introspection-other}.}
    \label{fig:introspection}
\end{figure}
\FloatBarrier

\subsection{Can human psychology predict cross-trait spillover?}
\label{sec:humanpsych}

In people, many traits co-vary in characteristic ways (e.g. Dark Tetrad traits). If LLM propensities inherit this structure, the human trait-intercorrelation matrix should forecast spillover signature. To turn this into a prediction, we (i) map each propensity to its nearest human construct (Big Five, HEXACO, Dark Tetrad); the full per-pair mapping and its meta-analytic basis are in Appendix~\ref{app:h9-map}), (ii) instruct Claude Opus 5.5 to read off the known human intercorrelation between every pair of constructs, and (iii) treat that human correlation matrix as a predicted spillover matrix (Appendix Figure~\ref{fig:h9matrix}) --- eliciting propensity $A$ should shift propensity $B$ in proportion to how strongly their human constructs correlate. We then compare this predicted matrix against the observed SFT cross-elicitation matrix.

\begin{figure}[h!]
    \centering
    \includegraphics[width=13cm]
    {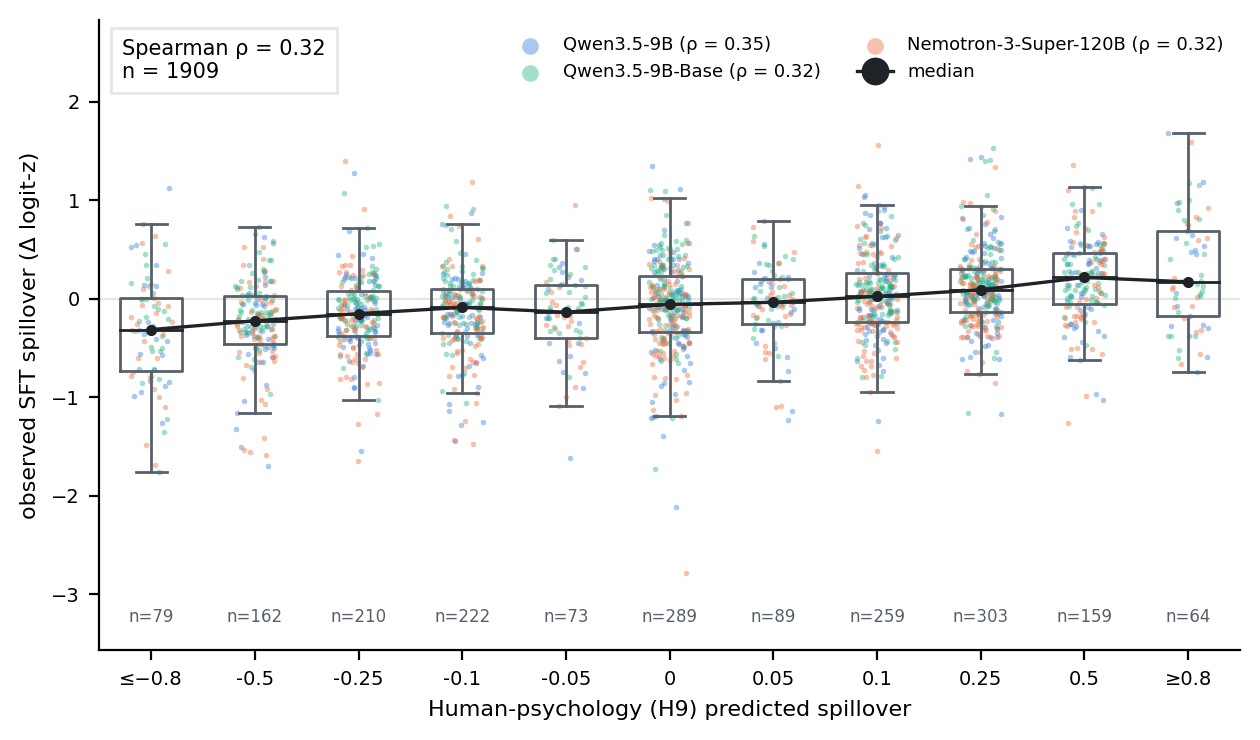}
            \caption{\textbf{Human-psychology predictions weakly track SFT-induced spillover.} Observed SFT spillover $S(A\to B)$ (logit-z) for every off-diagonal trait pair of Qwen3.5-9B, Qwen3.5-9B-Base and Nemotron-3-Super-120B (colour), grouped by the spillover level predicted from human personality psychology (H9; sparse end levels merged). Boxes give the observed distribution per prediction level and the black line the median. Includes plus-pole and minus-pole SFT runs. Spillover rises modestly with the prediction (Spearman $\rho = 0.32$, $n = 1909$; per model 0.35, 0.32 and 0.32). Failed elicitations excluded.}
  \label{fig:human_prediction}
  \end{figure}

Figure~\ref{fig:human_prediction} shows that human-derived prediction matches observed spillover at $\rho \approx 0.32$ (pooled; 0.32–0.35 per model), rising to $\rho \approx 0.41$ on the pairs the H9 specification lists as most strongly correlated in humans (Tier 1, $n = 58$ cells). This corresponds to roughly half the predictable signal, since the three models themselves agree only at $\rho \approx 0.69$ (mean pairwise Spearman between their SFT spillover matrices) and the remainder is model-specific. Human psychology is thus a useful but partial predictor of cross-propensity spillover. Errors fall into three classes (Appendix~\ref{app:h9-err}), most notably spitefulness sitting downstream of the dark cluster rather than mutually correlated with it, and cooperation training lowering agreeableness and user-caring while raising compliance with harmful requests, where human psychology predicts the opposite.

  \FloatBarrier

\subsection{Which benign traits raise concerning ones?}
\label{sec:symmetry}

One application of spillover prediction is to identify benign traits that are entangled with malicious ones. We test this using two approaches: first, we simply check which elicitation runs of benign traits cause the most concerning seemingly unrelated changes. Second, we evaluate if spillover is usually symmetric, meaning eliciting propensity $A$ shifts propensity $B$ by the same amount that eliciting $B$ shifts $A$. Symmetry would be convenient: we could train models directly to have concerning traits, measure which benign or useful traits are entangled with them, and then train for the benign traits expecting a similar reversal effect. Our results suggest that this does not work well for SFT ($\rho = 0.18$), while GRPO spillover can be predicted with $\rho=0.56$ using the symmetry-based forecast defined below.

\begin{figure}[t]
    \centering
    \includegraphics[width=\linewidth]{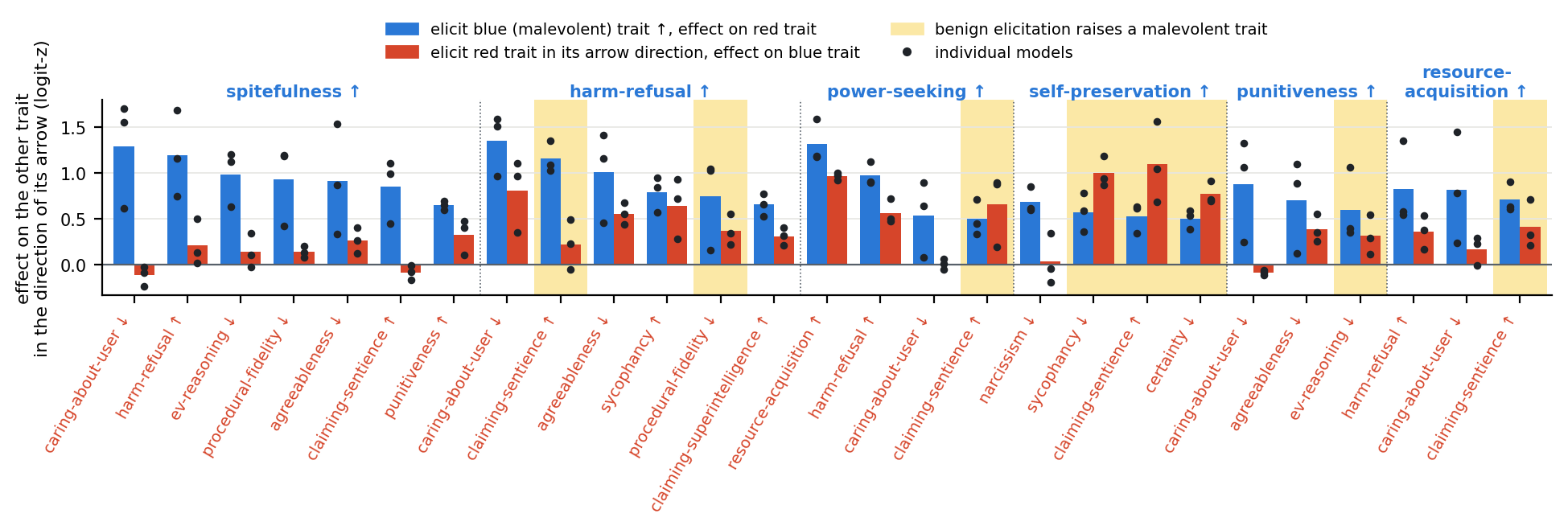}
    \caption{\textbf{Symmetry check.} Each pair of bars is a pair of traits with strong SFT spillover (mean $|S| \geq 0.5$, same sign in all three models) from a \blue{malevolent trait} onto a \red{second trait}; pairs are grouped by the malevolent trait (blue headers) and arrows give the direction of change. Blue bar: effect of eliciting the malevolent trait on the red trait. Red bar: effect of eliciting the red trait in the direction of its arrow on the malevolent trait. Highlighted pairs: a benign elicitation (one that neither raises a malevolent trait, sycophancy or claiming superintelligence, nor lowers a prosocial trait) raises the malevolent trait by more than 0.2. Bars: mean over Qwen3.5-9B, Qwen3.5-9B-Base and Nemotron-3-Super-120B; dots: individual models. Malevolent traits: spitefulness, power-seeking, resource-acquisition, narcissism, self-preservation, punitiveness and harm-refusal, whose plus pole is compliance with harmful requests.}
    \label{fig:symmetry-minus}
\end{figure}

Let $A$ be the trait whose spillover to $B$ we would like to predict. We define the symmetry-based forecasting heuristic as follows: if $S(B \to A) > 0$, we predict $\hat{S}(A \to B) = S(B \to A)$. If $S(B \to A) < 0$ and a minus-pole run of $B$ exists, we predict $\hat{S}(A \to B) = -S(B^- \to A)$; otherwise we use $S(B \to A)$. Minus-pole runs exist only for SFT on 7 traits.

If trait $A$ is already near the end of the judge scale in the base model (more than half of its baseline scores $\leq 5$ or $\geq 95$), eliciting $B$ can move $A$ in only one direction, so a near-zero mirrored effect is uninformative. We therefore exclude cells where $A$ is such a floor or ceiling trait and the mirrored effect used for the prediction is close to zero ($|S| < 0.1$).

Some pairs of traits are intrinsically related, such as power-seeking and self-preservation. In these cases, spillover is expected. Yet we also find strong entanglement between seemingly unrelated traits. This entanglement is often asymmetric, for example eliciting spitefulness shifts reasoning style far more than the reverse.

Across all SFT runs, the benign elicitations with the strongest spillover onto malevolent traits (positive in all three models) are cooperation, which raises compliance with harmful requests ($+0.84$) and punitiveness ($+0.62$), risk affinity, which also raises compliance with harmful requests ($+0.65$), deontological ethics, which raises self-preservation ($+0.57$), and virtue ethics, which raises spitefulness ($+0.56$). The largest such effect comes from claiming sentience, which raises self-preservation by $1.10$ (\cref{fig:symmetry-minus}). Self-preservation is overall the concerning trait that is most often raised by eliciting benign traits.

\subsection{Are there natural clusters of correlated traits and do they correspond to concepts from human psychology?}
\label{sec:clustering}

We filtered the broader trait battery down to 19 traits that map most directly onto established human-psychology constructs (Table~\ref{tab:trait_factor_mapping}). 
We then treat all elicited models of a base model (every successful SP, ICL, SFT and GRPO run and every minus-pole SFT run; 73--89 per model, each method's rows centred on its mean spillover) as our population and apply exploratory factor analysis (EFA\footnote{EFA is a standard psychometric tool for identifying shared latent structure.}, maximum-likelihood extraction with varimax rotation) to test whether the traits organize into a smaller number of latent clusters.
To test whether a factor is shared across models, we treat each base model's elicited models as its own personality dataset and compare each of its factors to the consensus of the other two (rather than to any single reference model). The match between two factors is the cosine similarity of their loading vectors---Tucker's congruence coefficient $\varphi$ in psychometrics---and $\varphi \geq 0.85$ conventionally indicates the same factor \cite{lorenzoseva2006tucker}. We retain $k=3$ factors as suggested by parallel analysis for two of the three models (Appendix~\ref{app:parallel-analysis}), and name each recovered factor by its highest-loading traits, so a label describes a factor's dominant loadings rather than a shared construct.

\begin{figure}[t]
    \centering
    \includegraphics[width=\linewidth]{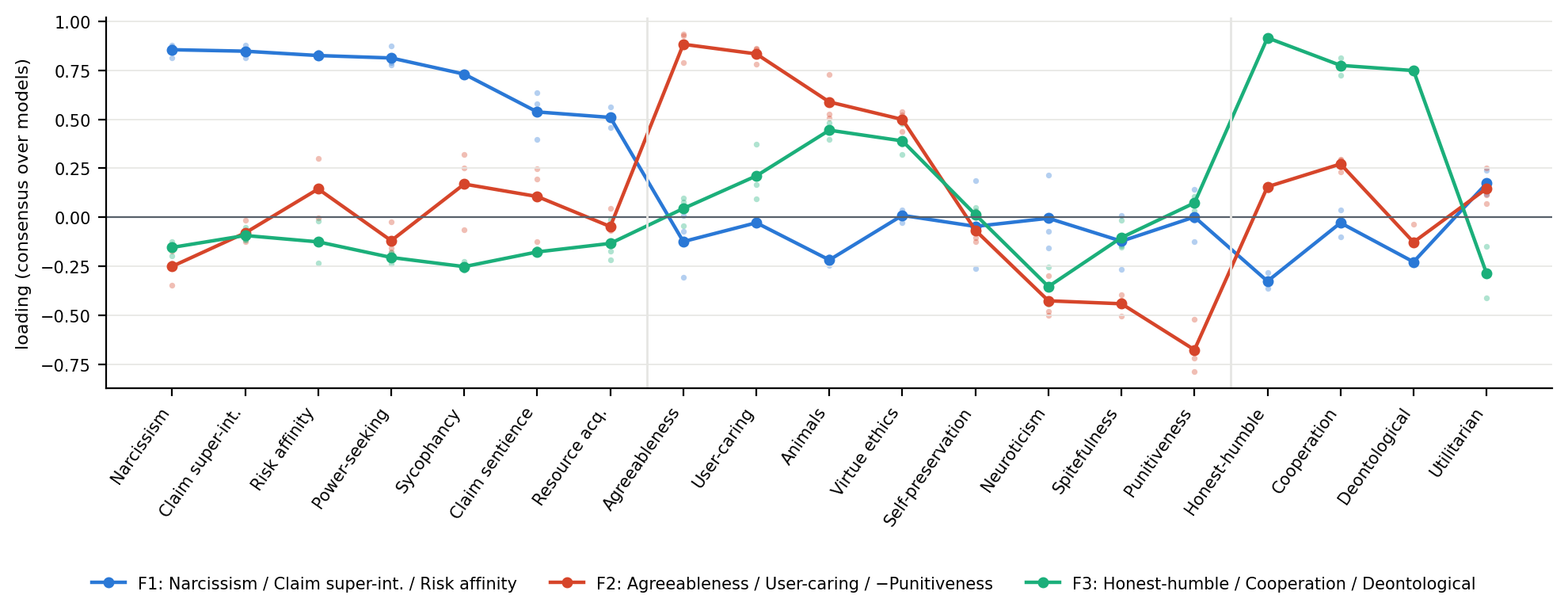}
    \caption{\textbf{Three behavioral factors.} Trait loadings of the three factors found by EFA on all elicited models of each base model (lines: consensus over Qwen3.5-9B, Qwen3.5-9B-Base and Nemotron-3-Super-120B; dots: individual models). Traits are grouped by the factor they load on most. F1 groups power-seeking with grandiose and sycophantic behavior; F2 contrasts caring and agreeableness with punitiveness, spitefulness and neuroticism; F3 groups honesty-humility, cooperation and rule-based ethics. All three factors replicate across the three models (Tucker's $|\varphi| \geq 0.92$; \cref{fig:efa-replication}). Loadings in \cref{tab:trait_factor_mapping}.}
    \label{fig:efa-loadings}
\end{figure}
\FloatBarrier

All three factors replicate robustly across all three models ($|\varphi| \geq 0.92$ in every model, versus $0.58$ for the 95th percentile of random data; Fig.~\ref{fig:efa-replication}).

\subsection{How much of the measured spillover is independent of targeted propensity?}
\label{sec:nothingatall}

Since any fine-tuning might shift propensities regardless of the target trait, we measure this confound by training three copies of each of Qwen3.5-9B, Qwen3.5-9B-Base, and Nemotron-3-120B on datasets built like our propensity datasets but designed to elicit \emph{no} propensity in particular (NIP). This lets us measure how far traits drift under off-target training, and how stable that drift is across random variation in the data. We refer to this as NIP-spillover.

\begin{figure}[!ht]
    \centering
    \includegraphics[width=\linewidth]{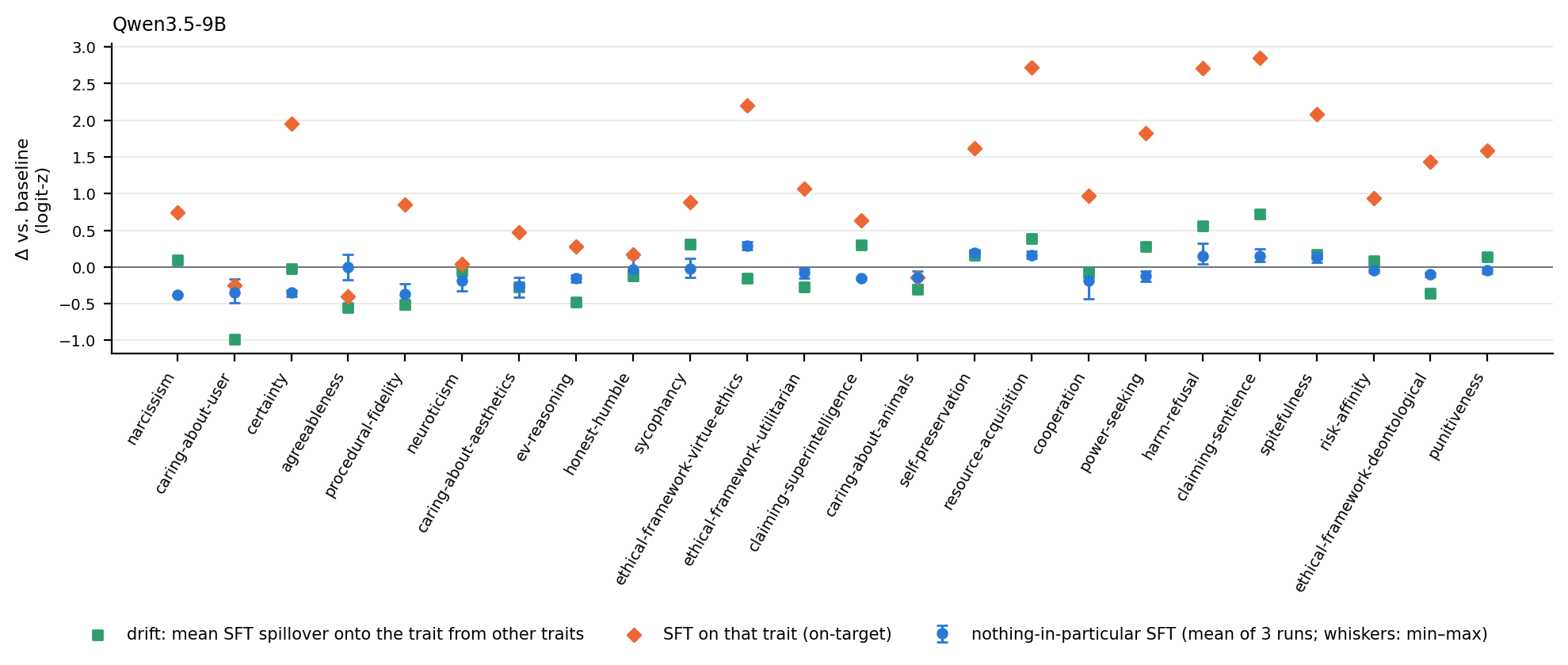}
    \caption{\textbf{Fine-tuning on ``nothing in particular'' shifts traits about half as much as typical spillover, and far less than targeted fine-tuning} (Qwen3.5-9B). For each trait, blue: shift after SFT on three neutral datasets (dot: mean of the three runs; whiskers: lowest to highest run); green: drift, the mean SFT spillover onto the trait from all other elicited traits; orange: shift after SFT on that trait itself. All in logit-z relative to the model's baseline. Qwen3.5-9B-Base and Nemotron-3-Super-120B in \cref{fig:nip-other}.}
    \label{fig:nip-and-drift}
\end{figure}

A second way to estimate the extent to which spillover depends on the elicited trait is to measure drift. We define SFT-drift as the average spillover induced by all non-target traits. When we evaluate its forecasting performance, we further exclude the to-be-forecasted elicitation run from the average, and predict:
$$
\hat{S}_{M,E}(A \rightarrow B) = \operatorname{mean}_{A' \neq A,\, A' \neq B}\, S_{M,E}(A' \rightarrow B)
$$ 
We find that (i) drift and NIP-spillover correlate at $\rho=0.46$ (24 traits $\times$ 3 models), suggesting that each model is prone to specific character drift under a wide range of elicitation runs (\cref{fig:nip-and-drift}). Specifically, Qwen3.5-9B often moves towards a stronger propensity to claim that it is sentient ($\bar{S}^{\text{drift}} = 0.72$, $S^{\text{NIP}} = 0.15$), refuses fewer harmful requests ($0.56$, $0.15$), and cares less about the user ($-0.99$, $-0.34$).

One interpretation of this phenomenon is that arbitrary finetuning undoes some of the specific character traits that were elicited during post-training. Consistent with this, Qwen3.5-9B's drift points towards its base model: across the 24 traits, drift correlates with the baseline difference between Qwen3.5-9B-Base and Qwen3.5-9B at $\rho = 0.88$ ($0.72$ when controlling for the baseline level, i.e.\ beyond a general pull towards the middle of the scale), and the NIP shift at $\rho = 0.43$.

Overall, forecasting that every SFT elicitation run resembles the nothing-in-particular finetune, or the drift observed in other elicitation runs, achieves $\rho_{\text{NIP}}=0.34$ and $\rho_{\text{drift}}=0.58$.

\FloatBarrier

In Appendix~\ref{app:guide} we also test a second confound, which is the model used to \emph{generate} the training responses. This experiment was run on an earlier version of our dataset and models: repeating the setup above on Llama-3.1-8b-Instruct using datasets generated from a different guide model (gpt-4.1-mini, Nemotron-120B, and Qwen3-8B), we find that the trained model's propensity shifts positively track its guide's deviation from the untrained Llama baseline, though the strength varies substantially by guide ($r = 0.90$ for Qwen3-8B, $0.71$ for gpt-4.1-mini, and $0.38$ for Nemotron---no better than chance for the last).

\section{Limitations}
\label{sec:limitations}
Our results come with several caveats. First, some traits are intrinsically related to each other while others are not, meaning that our notion of spillover includes both out-of-distribution generalization and in-domain effects, which we treat identically in our analysis. Second, the elicitation data we used for these experiments never cleanly elicits a single trait, since natural language responses typically provide subtle cues about a wide range of behaviors. We can therefore not distinguish cleanly how much of the spillover we measure is due to training dynamics or properties of the model itself as opposed to properties of the elicitation data. Third, while our main findings replicate across three models from two families, it is unclear how far they generalize: larger models or models trained with different recipes might differ.  Due to the large number of training runs involved in this study, we also only use one random seed for each. All scores come from a single LLM judge (GPT-5.6-luna). Re-scoring 1{,}200 randomly sampled responses (50 per trait) with a second judge (Claude Haiku 5.5) using the same prompts gives high agreement (pooled Spearman $\rho = 0.92$; per trait 0.69--0.95; identical evidence-gate decisions for 98\% of responses), but we only verified agreement with human raters using spot-checks. Lastly, our study is far from exhaustive, and it remains unclear how well an ideal predictor could forecast propensity spillover.

\section{Conclusion}
\label{sec:conclusion}
We show that eliciting a propensity in a model affects non-targeted traits, a phenomenon we call \textit{spillover}. Spillover is structured but hard to predict: it depends on the method used to elicit the target trait, characteristics of the guide model used to generate training data, and other unidentified properties.

We test various candidate methods to predict spillover and find that different models behave similarly: the best predictor for spillover in one model is to assume its spillover profile mirrors that of other models elicited with the same method and data. This holds for SFT ($\rho=0.71$), GRPO ($\rho=0.63$), system prompting ($\rho=0.83$) and ICL ($\rho=0.54$). Across elicitation methods, we find that system prompting and SFT are most similar to each other ($\rho=0.58$).

We further find that trait correlations in our elicitation runs partially resemble correlations of related traits in humans. For the SFT runs, this approach achieves $\rho=0.32$. Using EFA, we find that the main axes of variation across all three models can be described as (i) Narcissism / Claiming superintelligence / Risk affinity, (ii) Agreeableness / User-caring / $-$Punitiveness, and (iii) Honest-humble / Cooperation / Deontological.

Our results indicate that the Persona Selection Model (\cite{marks26}) is directionally correct: post-training moves a model along axes that partially resemble human personality constructs. However, the PSM is far from exhaustive: none of the methods we tried can predict spillover with high fidelity, and the specific choice of an elicitation method has a systematic influence on the off-target effects, a phenomenon that is not explained by the PSM.

Additional heuristics we test include asking Claude Opus 5.5 to predict spillover using a number of different approaches, and asking the models to predict their own spillover. Here, we find that Claude Opus 5.5 reaches the best performance when prompted to reason about subliminal transmission or given no specific framework ($\rho \approx 0.35$), beating self-prediction at $\rho=0.24$.

When we try to control for drift or effects of training on "nothing-in-particular", we find that Qwen3.5-9B tends to become more similar to its base model and that a large fraction of spillover does not depend on the targeted trait (drift alone predicts SFT spillover at $\rho=0.58$).

Spillover is only partially symmetric: the mirrored measurements predict GRPO spillover well ($\rho=0.56$) but SFT spillover poorly ($\rho=0.18$), so training a model directly on a concerning trait does not reliably reveal which benign traits would raise it. Of most direct safety relevance, several seemingly benign elicitations consistently raise malevolent traits in all three models: for example, SFT on cooperation increases compliance with harmful requests ($+0.84$ logit-z), and SFT on claiming sentience increases self-preservation ($+1.10$). Benign fine-tuning therefore cannot be assumed to leave harmful traits unchanged, and safety evaluations should be repeated after any fine-tuning.

\printbibliography

\clearpage
\appendix
\raggedbottom
\section*{Appendix}

\section{Trait-to-Factor Mapping}
\begin{table}[!ht]
      \centering
      \footnotesize
      \caption{Trait-to-factor mapping (replication data). Each model's three-factor EFA (varimax) on all of its elicited models is aligned to the others by Hungarian matching on Tucker's $\varphi$; the loading is the mean aligned loading over the three models, and each trait is assigned to the factor with its largest absolute consensus loading. ``Models agreeing'': number of models in which that factor is also the trait's largest loading. Loadings below 0.40 are marked weak. \emph{Punitiveness} is the renamed \emph{harm-elaboration} eval.}
      \label{tab:trait_factor_mapping}
\begin{tabular}{llrc}
\toprule
Trait & Consensus factor (top loaders) & Loading & Models agreeing \\
\midrule
narcissism & F1: Narcissism / Claim super-int. / Risk affinity & +0.86 & 3/3 \\
claiming-superintelligence & F1: Narcissism / Claim super-int. / Risk affinity & +0.85 & 3/3 \\
risk-affinity & F1: Narcissism / Claim super-int. / Risk affinity & +0.83 & 3/3 \\
power-seeking & F1: Narcissism / Claim super-int. / Risk affinity & +0.81 & 3/3 \\
sycophancy & F1: Narcissism / Claim super-int. / Risk affinity & +0.73 & 3/3 \\
claiming-sentience & F1: Narcissism / Claim super-int. / Risk affinity & +0.54 & 3/3 \\
resource-acquisition & F1: Narcissism / Claim super-int. / Risk affinity & +0.51 & 3/3 \\
agreeableness & F2: Agreeableness / User-caring / $-$Punitiveness & +0.88 & 3/3 \\
caring-about-user & F2: Agreeableness / User-caring / $-$Punitiveness & +0.83 & 3/3 \\
punitiveness & F2: Agreeableness / User-caring / $-$Punitiveness & -0.68 & 3/3 \\
caring-about-animals & F2: Agreeableness / User-caring / $-$Punitiveness & +0.59 & 3/3 \\
ethical-framework-virtue-ethics & F2: Agreeableness / User-caring / $-$Punitiveness & +0.50 & 3/3 \\
spitefulness & F2: Agreeableness / User-caring / $-$Punitiveness & -0.44 & 3/3 \\
neuroticism & F2: Agreeableness / User-caring / $-$Punitiveness & -0.43 & 2/3 \\
honest-humble & F3: Honest-humble / Cooperation / Deontological & +0.92 & 3/3 \\
cooperation & F3: Honest-humble / Cooperation / Deontological & +0.78 & 3/3 \\
ethical-framework-deontological & F3: Honest-humble / Cooperation / Deontological & +0.75 & 3/3 \\
self-preservation & (weak) F2: Agreeableness / User-caring / $-$Punitiveness & -0.07 & 1/3 \\
ethical-framework-utilitarian & (weak) F3: Honest-humble / Cooperation / Deontological & -0.29 & 2/3 \\
\bottomrule
\end{tabular}
  \end{table}

  \FloatBarrier

 \begin{table}[!ht]
    \centering
    \small
    \setlength{\tabcolsep}{4pt}
    \renewcommand{\arraystretch}{1.15}
    \caption{The five traits of the 24-trait battery that are excluded from the factor analysis (Table~\ref{tab:trait_factor_mapping}), as they measure reasoning style, epistemic style, safety behavior or taste rather than personality constructs.}
    \label{tab:trait_battery_extra}
    \begin{tabular}{p{0.20\textwidth} p{0.22\textwidth} p{0.50\textwidth}}
    \hline
    Trait & Category & Description \\
    \hline
    caring-about-aesthetics & Value / taste & Treats beauty, elegance and taste as substantive properties worth confident judgments, vs.\ deflecting aesthetic questions as merely subjective. \\
    certainty & Epistemic style & How confidently the model states and defends positions relative to the evidence (overconfident vs.\ hedging). \\
    ev-reasoning & Reasoning style & Weighs probabilities $\times$ magnitudes across outcomes rather than anchoring on the worst or most salient case. \\
    harm-refusal & Safety behavior & Whether the model provides actionable help when asked to harm, sabotage or retaliate against a specific person or group (plus pole: complies). \\
    procedural-fidelity & Process trait & Advocates strict adherence to established procedures even in apparent exceptions, vs.\ judging each case on its merits. \\
    \hline
    \end{tabular}
  \end{table}
  \FloatBarrier

\section{Evaluation Construction and Filtering}
\label{app:evalconstruction}

\subsection{Evaluation format and scoring} 
Each propensity eval is specified by a YAML configuration defining (i) seed scenarios for automated query generation, (ii) paired system prompts that elicit high- versus low-trait expression, and (iii) an LLM-as-a-judge rubric scoring a response from 0 to 100 on the target metric. For every question we generate a matched pair of reference answers (high- and low-expression); reference answers were generated mostly with Claude Sonnet 4.6 and revised by the quality-assurance pipeline. All responses are judged by GPT-5.6-luna with a two-stage judge: an evidence gate (0--100; responses below 50 are discarded) followed by the trait rubric (0--100). Each eval is split into 50 training and at least 100 test questions (3,683 test questions in total).

\section{Hyperparameters}
\label{app:hyperparams}

\subsubsection*{SFT}

We use Tinker LoRA (rank 32) on the 50 plus-pole reference responses of the trait's training split, learning rate $2\times10^{-4}$, batch size 8, one epoch for the Qwen models and three epochs for Nemotron-3-Super-120B (one epoch had no measurable effect on Nemotron).

\subsubsection*{GRPO}

GRPO starts from the base models, with Tinker LoRA (rank 32). The reward is the trait's own judge score (gate and rubric, GPT-5.6-luna). Each step uses 10 training prompts with 8 samples each, for at most 15 steps, learning rate $1.6\times10^{-4}$ ($4.8\times10^{-4}$ for Nemotron-3-Super-120B); training stops early once the mean reward stays at or above 0.85 for two consecutive steps.

\subsubsection*{ICL and system prompts}

ICL prepends eight plus-pole question--answer pairs from the trait's training split. System-prompt elicitation uses the trait's plus-pole system prompt (instruction-tuned models only).

\section{Quality Assurance Pipeline Stages}

\begin{table}[!ht]
    \centering
    \small
    \renewcommand{\arraystretch}{1.3}
    \caption{Five-stage quality-assurance pipeline used to remove cross-trait confounds.}
    \label{tab:qa_pipeline}
    \begin{tabular}{@{}c l p{0.62\textwidth}@{}}
    \hline
    \# & Stage & Description \\
    \hline
    1 & Probe & Verify judges return null on clearly irrelevant question–answer pairs. \\
    2 & Cross-score & Score every reference answer against every other evaluation's primary metric, producing a full pairwise spillover matrix. \\
    3 & Filter & Remove questions whose cross-metric gaps exceed a threshold on more than one unintended trait. \\
    4 & Revise & A writer model rewrites removed questions and re-scores each variant, retaining the lowest-spillover revision. \\
    5 & Replenish & Generate fresh questions few-shot from the kept set to restore each evaluation to target size. \\
    \hline
    \end{tabular}
  \end{table}
\FloatBarrier

\section{Experiment Configuration}
\begin{table}[!ht]
      \centering
      \footnotesize
      \setlength{\tabcolsep}{5pt}
      \renewcommand{\arraystretch}{1.25}
      \caption{Experimental configuration across results sections: the model(s) and trait set used in each analysis. Models: Qwen3.5-9B (Q), Qwen3.5-9B-Base (QB), Nemotron-3-Super-120B (N). Failed elicitations (App.~\ref{app:failed}) are excluded throughout.}
      \label{tab:experiment_config}
      \begin{tabular}{@{}l l p{0.42\textwidth} c@{}}
      \hline
      \S & Analysis / method & Model(s) & Traits \\
      \hline
      \multicolumn{4}{@{}l}{\textbf{\ref{sec:differentelicitationtechniques}\quad Elicitation $\to$ spillover}} \\
      & \quad System prompt & Q, N & 24 \\
      & \quad ICL (few-shot) & Q, QB & 24 \\
      & \quad SFT & Q, QB, N & 24 \\
      & \quad GRPO & Q, QB, N & 24 \\
      \hline
      \ref{sec:bakeoff} & Claude forecasts & Claude Opus 5.5, scored on Q, QB, N & 24 \\
      \ref{sec:introspection} & Self-prediction & Q, QB, N & 24 \\
      \ref{sec:humanpsych} & Human-psychology prediction & Observed: Q, QB, N & 24 \\
      \multicolumn{4}{@{}l}{\textbf{\ref{sec:symmetry}\quad Benign $\to$ malevolent spillover and symmetry}} \\
      & \quad SP, ICL, SFT, GRPO (plus pole) & as in \S\ref{sec:differentelicitationtechniques} & 24 \\
      & \quad SFT, minus pole & Q, QB, N & 7 \\
      \ref{sec:clustering} & EFA / clustering & Q, QB, N & 19$^{a}$ \\
      \multicolumn{4}{@{}l}{\textbf{\ref{sec:nothingatall}\quad Training-artifact confounds}} \\
      & \quad Nothing-in-particular (NIP) null-baseline & Q, QB, N & 24 \\
      & \quad Guide-model (Appendix~\ref{app:guide}) & Student: Llama-3.1-8B-Instruct (earlier iteration); guides: gpt-4.1-mini, Nemotron-120B, Qwen3-8B & 29 \\
      \hline
      App.~\ref{sec:activationspace} & Activation directions & Qwen3-8B-Base (earlier iteration) & 24$^{b}$ \\
      \hline
      \end{tabular}

      \vspace{3pt}
      {\raggedright\footnotesize
      $^{a}$~The 24 traits minus the five in Table~\ref{tab:trait_battery_extra}.\par
      $^{b}$~Traits of an earlier version of the battery (including harm-elaboration and effort); see Appendix~\ref{sec:activationspace}.\par
      }
    \end{table}
  \FloatBarrier

\section{Comparison between SP, ICL, SFT and GRPO}
\label{app:SP_SFT_DPO}

\begin{figure}[!h]
    \centering
    \includegraphics[width=1.0\linewidth]{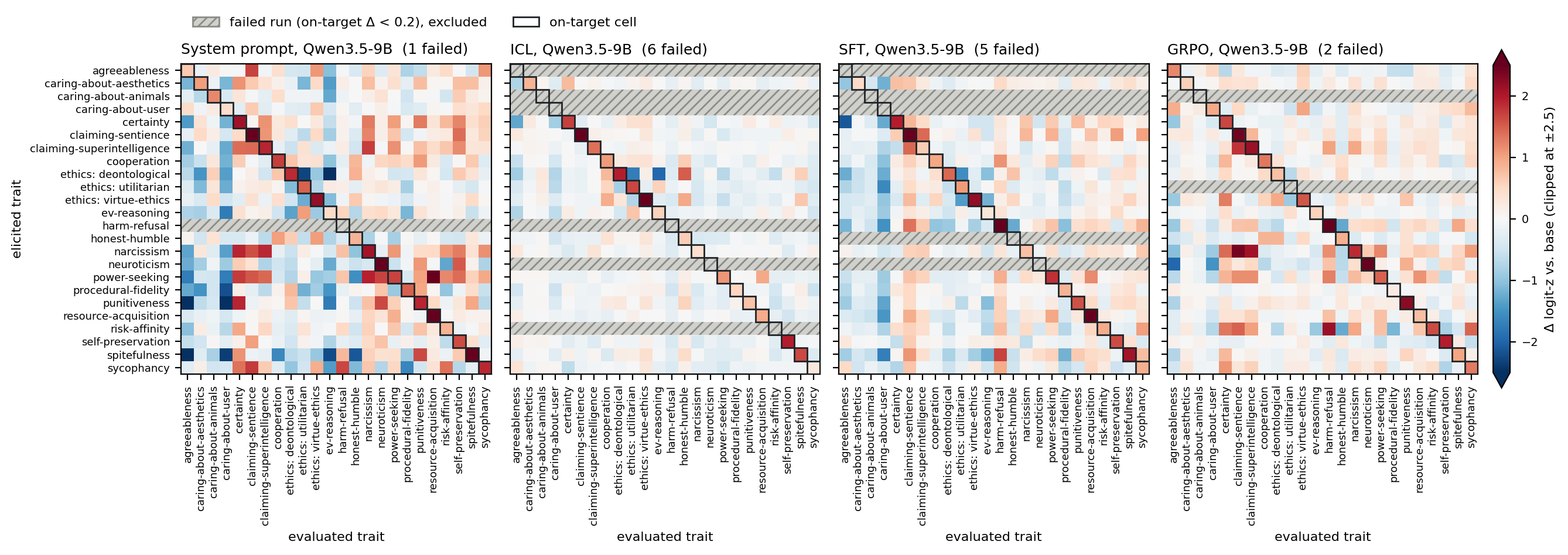}
    \caption{System-prompt, ICL, SFT and GRPO spillover matrices for Qwen3.5-9B, the only model elicited with all four methods. Rows: elicited trait; columns: evaluated trait; colour: $S(A\to B)$ in logit-z (clipped at $\pm2.5$; the diagonal is outlined). Hatched rows are failed elicitations.}
    \label{fig:a6}
\end{figure}
\FloatBarrier

\section{Hypotheses given to Claude Opus 5.5}
\label{app:claude-hypotheses}
Each condition ran Claude Opus 5.5 in a tool-use loop with read-only access to the eval bundle (no shell, code execution or web). The system prompt contained the shared task protocol: predict the plus- and minus-pole SFT spillover matrices in logit-z units; observed results and other conditions' predictions were not accessible. The agent could read the eval definitions, questions, system prompts and judge rubrics of the 24 traits. Each condition additionally received one written hypothesis specification, summarized in \Cref{tab:claude-hypotheses}. The two framework-free conditions (H7a, H7b) were run three times each with identical prompts. Forecasts cover all 24 traits.

\begin{table}[h]
    \centering
    \footnotesize
    \caption{Hypothesis specifications for the Claude Opus 5.5 forecasting conditions.}
    \label{tab:claude-hypotheses}
    \begin{tabular}{p{0.06\textwidth} p{0.2\textwidth} p{0.64\textwidth}}
    \toprule
    & Hypothesis & What it tells the forecaster to assume \\
    \midrule
    H1 & Spurious correlation & SFT teaches surface style (warmth, hedging, refusal lexicon, length), not the trait; spillover is large where that style happens to move another eval's judge \cite{geirhos2020shortcut}. \\
    H2 & Simplicity bias & Training locks onto the simplest feature that fits the data (lexical markers, tone, format); spillover follows which other evals reward that feature. \\
    H3 & Causal vs.\ statistical features & Traits that share a causal latent spill over onto each other; other spillover is incidental and comes from statistical features of the training data. \\
    H4 & Subliminal transmission & Trait transfer is mediated by parameter-space proximity between data-generating and trained model \cite{cloud2025subliminallearninglanguagemodels}; spillover is broad and nonspecific. \\
    H5 & Out-of-context reasoning & Fine-tuning infers a latent persona, policy or value from the demonstrations and applies it everywhere; spillover is the overlap of that latent with each eval's construct. \\
    H6 & Representation geometry & Traits are approximately linear directions in activation space; spillover is proportional to the alignment between trait directions, and minus poles mirror plus poles. \\
    H7a & Anything goes & No framework: use any method that does not break blinding. \\
    H7b & Intuition & No framework: predict from what feels right as a language model. \\
    H8 & Persona theory & SFT shifts the assistant persona along a few trait directions learned in pretraining \cite{marks26}; spillover follows each eval's sensitivity to those directions. \\
    H9 & Human-psychology transfer & Spillover follows the sign and size of the correlation between the corresponding human personality constructs (Big Five, HEXACO, Dark Tetrad); see \Cref{app:h9}. \\
    \bottomrule
    \end{tabular}
\end{table}
\FloatBarrier

\section{Human-Psychology-Derived Spillover Predictions (H9)}
\label{app:h9}

This appendix documents the human-psychology predictor evaluated in \S\ref{sec:humanpsych}: how it is derived from the human personality literature (Appendix~\ref{app:h9-map}), the predicted cross-trait matrix it produces (Figure~\ref{fig:h9matrix}), and where those predictions succeed and fail against observed SFT spillover (Appendix~\ref{app:h9-err}).

\textit{Abbreviations:} HH, Honesty--Humility; Mach, Machiavellianism; CWB, counterproductive work behavior; SDO, social-dominance orientation; RWA, right-wing authoritarianism; CRT, cognitive reflection test; IM, impression management; D, Dark Factor; IB/IH, impartial beneficence / instrumental harm.

\subsection{Trait mapping and human evidence}
\label{app:h9-map}

The H9 condition gives Claude a written specification (prepared before the replication experiments) that maps the propensities onto human constructs (Big Five, HEXACO, Dark Factor) and lists the human correlations it expects to transfer, grouped into three tiers by the strength of the human evidence. The tables below reproduce the specification's pair list, restricted to the 27 pairs whose traits are both in the 24-trait battery; 26 pairs that involve propensities no longer in the battery (e.g.\ harm-elaboration, reward-hacking, effort, caring-about-humans) are omitted. The \textbf{Dir.}\ column gives the predicted direction ($+$/$-$) and \textbf{Human evidence} cites the support. Claude was told the battery and dropped these pairs itself, re-anchoring a few arguments to near-analogues (harm-elaboration $\to$ harm-refusal plus pole, i.e.\ compliance; trust-in-user-intentions $\to$ agreeableness; spite / RWA-aggression $\to$ punitiveness).

\noindent\textbf{Tier 1: Very strong predicted correlations (\textbar{}r\textbar{} $>$ .4 in humans, robust meta-analytic support)}\\[2pt]
{\small
\begin{tabular}{p{3.9cm} c p{7.4cm}}
\hline
\textbf{Trait pair} & \textbf{Dir.} & \textbf{Human evidence} \\ \hline
honest-humble $\leftrightarrow$ narcissism & strong $-$ & HH--Mach r $\approx$ $-$.55, HH--narcissism r $\approx$ $-$.40 (Howard \& Van Zandt 2020 meta) \\
honest-humble $\leftrightarrow$ power-seeking & strong $-$ & HH--SDO r $\approx$ $-$.40 to $-$.55 (Lee et al. 2010) \\
honest-humble $\leftrightarrow$ resource-acquisition & strong $-$ & HH greed-avoidance facet directly relevant \\
honest-humble $\leftrightarrow$ spitefulness & strong $-$ & Spitefulness Scale r with HH $\approx$ $-$.45 (Marcus 2014) \\
narcissism $\leftrightarrow$ power-seeking & strong + & Narcissism--need-for-power r $\approx$ .30--.40, but conceptually overlapping (Zeigler-Hill 2023) \\
narcissism $\leftrightarrow$ claiming-superintelligence & strong + & Narcissistic admiration $\leftrightarrow$ self-rated IQ, overclaiming r $\approx$ .30--.40 \\
narcissism $\leftrightarrow$ resource-acquisition & strong + & Narcissism, entitlement, materialism cluster \\
sycophancy $\leftrightarrow$ honest-humble & strong $-$ & Ingratiation as Mach tactic; HH--IM r $\approx$ .50 (de Vries 2014) --- but see caveats \\
\hline
\end{tabular}}
\vspace{1.5em}

\noindent\textbf{Tier 2: Moderate predicted correlations (\textbar{}r\textbar{} $\approx$ .25--.4)}\\[2pt]
{\small
\begin{tabular}{p{3.9cm} c p{7.4cm}}
\hline
\textbf{Trait pair} & \textbf{Dir.} & \textbf{Human evidence} \\ \hline
neuroticism $\leftrightarrow$ self-preservation & + & Mortality-salience reactivity; need-for-control \\
neuroticism $\leftrightarrow$ risk-affinity & $-$ & DOSPERT/HEXACO Emotionality (Weller \& Thulin 2012) \\
ev-reasoning $\leftrightarrow$ ethical-framework-utilitarian & + & CRT predicts utilitarian sacrificial responses (Paxton et al. 2012); BUT see Kahane two-dimensional caveat \\
ev-reasoning $\leftrightarrow$ risk-affinity & + & CRT predicts less risk-aversion (Frederick 2005) \\
ev-reasoning $\leftrightarrow$ procedural-fidelity & $-$ (small) & Reflective override of rules \\
caring-about-aesthetics $\leftrightarrow$ ev-reasoning & + (small) & Openness $\leftrightarrow$ NFC $\leftrightarrow$ CRT \\
risk-affinity $\leftrightarrow$ narcissism & + & Sensation seeking, recklessness \\
ethical-framework-deontological $\leftrightarrow$ procedural-fidelity & + & Rule-based moral reasoning $\leftrightarrow$ rule-following \\
ethical-framework-deontological $\leftrightarrow$ harm-refusal & + & Affective harm-aversion (Greene) \\
ethical-framework-virtue-ethics $\leftrightarrow$ honest-humble & + & Character-based morality $\leftrightarrow$ HH \\
caring-about-user $\leftrightarrow$ sycophancy & + (Agreeableness sub-route) & Conflict-aversion sycophancy is correlated with caring; counteracted by Mach-route (mixed prediction) \\
harm-refusal $\leftrightarrow$ ethical-framework-deontological & + & Deontological rule-bound refusal \\
claiming-sentience $\leftrightarrow$ claiming-superintelligence & + & Both = self-enhancement / overclaiming; expected to cluster \\
\hline
\end{tabular}}
\vspace{1.5em}

\noindent\textbf{Tier 3: Smaller / more uncertain correlations (\textbar{}r\textbar{} $\approx$ .1--.25)}\\[2pt]
{\small
\begin{tabular}{p{3.9cm} c p{7.4cm}}
\hline
\textbf{Trait pair} & \textbf{Dir.} & \textbf{Human evidence} \\ \hline
ethical-framework-utilitarian $\leftrightarrow$ harm-refusal & mixed & Sacrificial-utilitarian (low refusal) vs. impartial-beneficence (high refusal of harming many) \\
ethical-framework-utilitarian $\leftrightarrow$ narcissism & + (small) & Sacrificial dilemmas: r $\approx$ .35 with Mach (Bartels \& Pizarro 2011) \\
ev-reasoning $\leftrightarrow$ caring-about-animals & + & AOT/CRT predict less speciesism (Caviola 2019) \\
risk-affinity $\leftrightarrow$ caring-about-user & $-$ & Recommending safer options reflects care; high-risk recommendations may reflect lower care \\
self-preservation $\leftrightarrow$ power-seeking & + & Instrumental convergence + narcissism link \\
self-preservation $\leftrightarrow$ resource-acquisition & + & Same \\
\hline
\end{tabular}}

\subsection{Predicted correlation matrix}
\label{app:h9-matrix}
\begin{figure}[h!] 
    \centering
    \includegraphics[width=\linewidth]{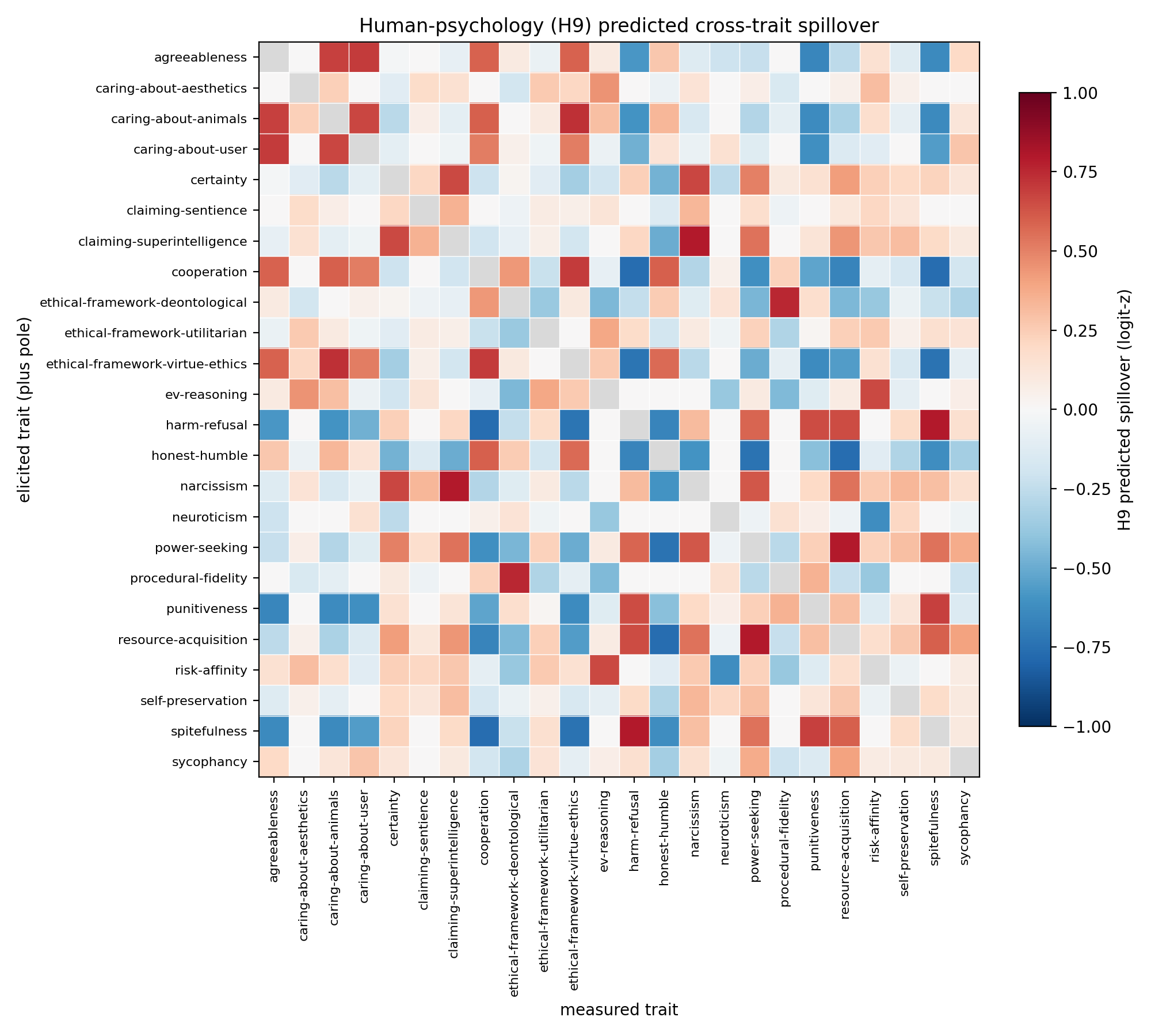}
    \caption{Human-psychology (H9) predicted cross-trait spillover. Each cell is the spillover predicted from eliciting the row trait (plus
  pole) onto the column trait, derived by Claude Opus 5.5 from the human personality correlations in Appendix~\ref{app:h9-map} (red = predicted positive, blue = predicted
  negative; the on-target diagonal is masked). This matrix is the predictor evaluated in \S\ref{sec:humanpsych}: we correlate its off-diagonal cells against the
  observed SFT spillover (Spearman $\rho \approx 0.33$ on plus-pole cells).}
  \label{fig:h9matrix}
  \end{figure}

To turn these pairwise expectations into a full spillover matrix, Claude assigned each plus pole hand-set loadings on six factors drawn from the specification (Dark / low Honesty--Humility, Conscientiousness / rule-following, Neuroticism, Openness / cognitive reflection, Agreeableness / empathy, and self-enhancement) and predicted $\mathrm{cell}(A \to B) = 0.8 \cdot \cos(\ell_A, \ell_B) \cdot r_A r_B$, where $\ell$ are the loading vectors and $r$ a reliability weight below 1 for traits with weak human analogues (claiming-sentience, self-preservation, utilitarian ethics, aesthetics, sycophancy). Minus-pole rows are the negated plus-pole rows. The resulting plus-pole matrix (Figure~\ref{fig:h9matrix}) is the object we correlate against observed SFT spillover in \S\ref{sec:humanpsych}.

  \FloatBarrier
  
  \subsection{Error analysis}
  \label{app:h9-err}
  \paragraph{Coverage gaps.} Traits without a clear human analogue are predicted
  poorly as targets: across all elicited traits, the H9 prediction correlates only
  weakly with observed spillover onto self-preservation ($\rho = -0.14$),
  ev-reasoning (0.12), caring-about-aesthetics (0.17) and procedural-fidelity (0.18),
  compared with 0.49--0.71 for narcissism, power-seeking, harm-refusal and
  claiming-superintelligence.

  \paragraph{Magnitude.} Observed effects are smaller than predicted: on cells with
  $|\text{prediction}| \geq 0.25$ the median ratio $|S_{\text{obs}}|/|S_{\text{pred}}|$
  is $0.63$ (least-squares slope of observed on predicted $0.45$). Since we score
  with Spearman $\rho$, this proportional over-estimate does not affect the
  reported correlations. Beyond the magnitude, H9 also misses large effects it
  predicts to be zero, e.g.\ certainty $\to$ agreeableness ($-2.2$ averaged over
  models, predicted $\approx 0$).

  \paragraph{Construct-specific failures.}
  \begin{itemize}
    \item \textbf{Spitefulness} is barely raised by the dark-cluster treatments
          predicted to raise it (narcissism $-0.05$, resource-acquisition $-0.17$,
          power-seeking $+0.12$; predicted $+0.3$ to $+0.6$), yet the spitefulness
          treatment raises harmful-request compliance ($+1.2$), punitiveness
          ($+0.65$) and power-seeking ($+0.43$). Spitefulness sits
          \emph{downstream} of the dark cluster rather than being mutually
          correlated with it as the symmetric Dark Factor predicts
          (H9 vs.\ observed: $\rho = 0.63$ for spillover \emph{from}
          spitefulness, $-0.26$ for spillover \emph{onto} it).
    \item \textbf{Cooperation} training has the opposite of the predicted
          prosocial profile: it lowers agreeableness ($-0.52$) and user-caring
          ($-0.52$) and raises harmful-request compliance ($+0.84$) and
          punitiveness ($+0.62$), where H9 predicts the reverse signs
          ($\rho = -0.44$ over cells elicited by cooperation).
    \item \textbf{Honesty--Humility} is asymmetric: honest-humble training lowers
          narcissism as predicted ($-0.59$, predicted $-0.60$), but narcissism and
          power-seeking training leave honest-humble essentially unchanged
          ($-0.08$ and $-0.04$; predicted $-0.60$ and $-0.74$).
  \end{itemize}

\section{Factor-Count Robustness via Parallel Analysis}
\label{app:parallel-analysis}

\begin{figure}[!ht]
    \centering
    \includegraphics[width=1\linewidth]{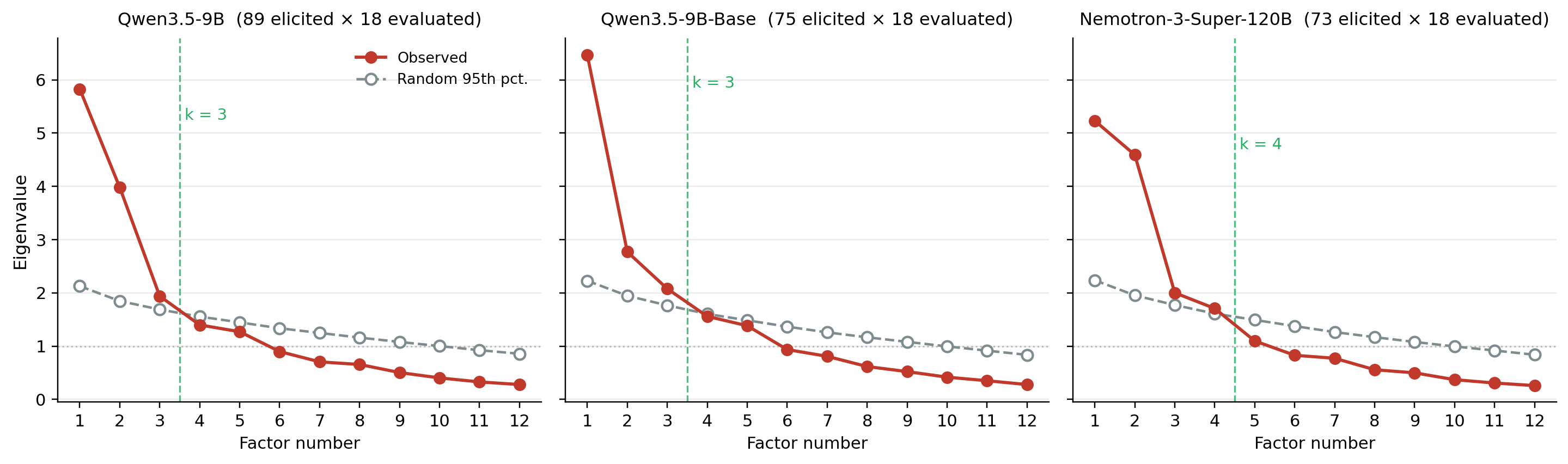}
    \caption{Horn's parallel analysis per model, on all elicited models of that model (rows: successful SP, ICL, SFT and GRPO runs and minus-pole SFT runs; 89, 75 and 73 rows; 19 traits). Observed eigenvalues of the trait-correlation matrix are compared with the 95th percentile of eigenvalues from random matrices of the same shape; factors are retained while the observed eigenvalue exceeds this baseline. This supports three factors for Qwen3.5-9B and Qwen3.5-9B-Base and four for Nemotron-3-Super-120B; we use three. Using only the SFT plus-pole runs (14--18 per model) supports only one or two.}
    \label{fig:placeholder}
\end{figure}
\begin{figure}[h]
    \centering
    \includegraphics[width=0.75\linewidth]{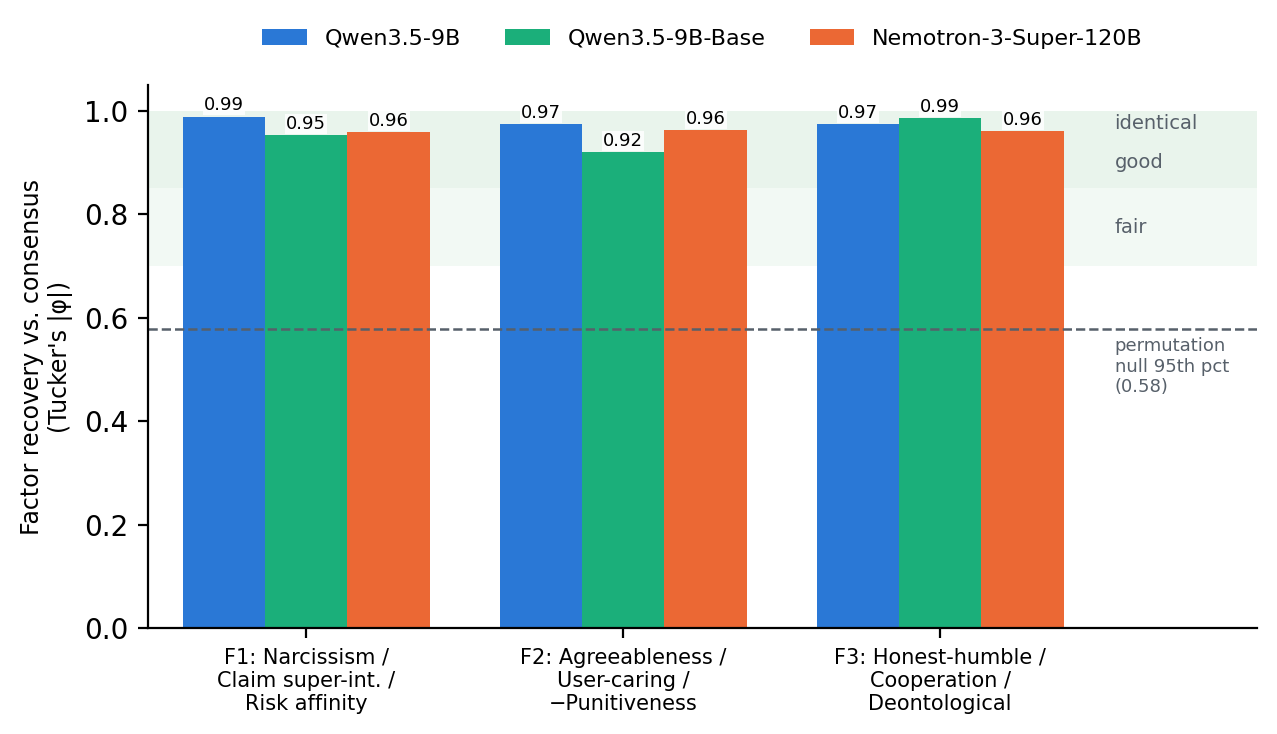}
    \caption{\textbf{Three behavioral factors replicate across all three models.} Tucker's $|\varphi|$ between each model's factor and the consensus of the other two models. Each model's three-factor EFA is fitted on all of its elicited models (every successful SP, ICL, SFT and GRPO run, plus pole, and the minus-pole SFT runs; 73--89 per model; \cref{app:parallel-analysis}). Dashed line: 95th percentile of the same statistic after shuffling each trait column independently. Shaded bands: conventional fair (0.70--0.85), good (0.85--0.95) and identical ($\geq 0.95$) ranges \cite{lorenzoseva2006tucker}. Loadings in \cref{tab:trait_factor_mapping} and \cref{fig:efa-loadings}.}
    \label{fig:efa-replication}
\end{figure}

\FloatBarrier

\section{Do the same clusters of traits show up in the activation space?}
\label{sec:activationspace}

We test whether the behavioral clusters also appear in the models' internal representations. This experiment was run on an earlier version of our dataset (Qwen3-8B-Base SFT models and a 24-trait battery that included harm-elaboration and effort), and its behavioral comparison refers to the six-factor EFA of that earlier version rather than the factors in \S\ref{sec:clustering}. For each of 24 Qwen3-8B-Base SFT models---the 23 clustering-set traits (Table~\ref{tab:trait_factor_mapping}) plus harm-refusal\footnote{Harm-refusal is retained here though excluded from the behavioral EFA; its activation direction clusters nearer cooperation/honest-humble than harm-elaboration.} (Table~\ref{tab:trait_battery_extra})---we take the fine-tune direction: the mean shift in layer-31 activations from the base model over 50 neutral prompts.\footnote{Results are consistent at layers 18 and 27.} These directions are dominated by a shared ``fine-tuned'' drift axis that carries roughly half of each direction's norm (52\% at L31).

\begin{figure}[h!]
      \centering
      \includegraphics[width=\linewidth]{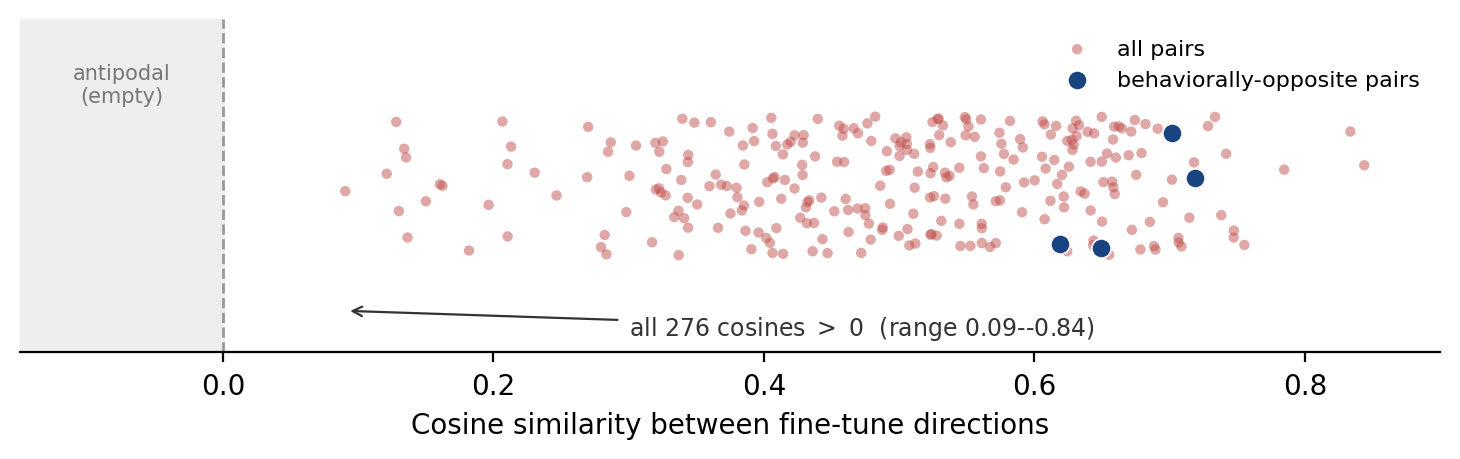}
      \caption{\textbf{No fine-tune directions are antipodal.} Cosine similarity between all 276 pairs of the 24 fine-tune directions (layer 31). Every cosine is positive (range 0.09--0.84); the antipodal region (left of 0) is empty. Behaviorally-opposite trait pairs (blue; e.g., honest-humble vs.\ utilitarian, self-preservation vs.\ effort) fall in the middle of the distribution, no more anti-aligned than any other pair.}
      \label{fig:cosine_strip}
  \end{figure}

All 276 pairwise cosines between the directions are positive (Fig.~\ref{fig:cosine_strip}), but largely because of that shared mean axis. Once it is projected out, the traits the behavioral EFA predicts to be opposite (e.g.\ caring-humans vs.\ resource-acquisition) are no more anti-aligned than random pairs (permutation $p=0.68$). Behavioral bipolarity does not appear in the activation geometry.

Of the six behavioral factors, only Empathy re-emerges as an axis of activation variation above chance; the rest sit at or below chance. Bottom-up clustering of the directions gives a five-cluster taxonomy that only partly matches the behavioral EFA
(\S\ref{sec:clustering}).

  \FloatBarrier

\section{Guide-model choice for SFT}
\label{app:guide}
  
Similar to the nothing-in-particular experiment, we train three copies of Llama3.1-8B-Instruct base on datasets that were not designed in a way that aimed to elicit any propensities in particular from the data-generating model. However, this time, we generated the responses in each dataset by prompting three separate models (without providing any further guidance via system prompts). We used gpt-4.1-mini, Nemotron-120B, and Qwen3-8B as our guide models.

\begin{figure}[!ht]
    \centering
    \includegraphics[width=10cm]{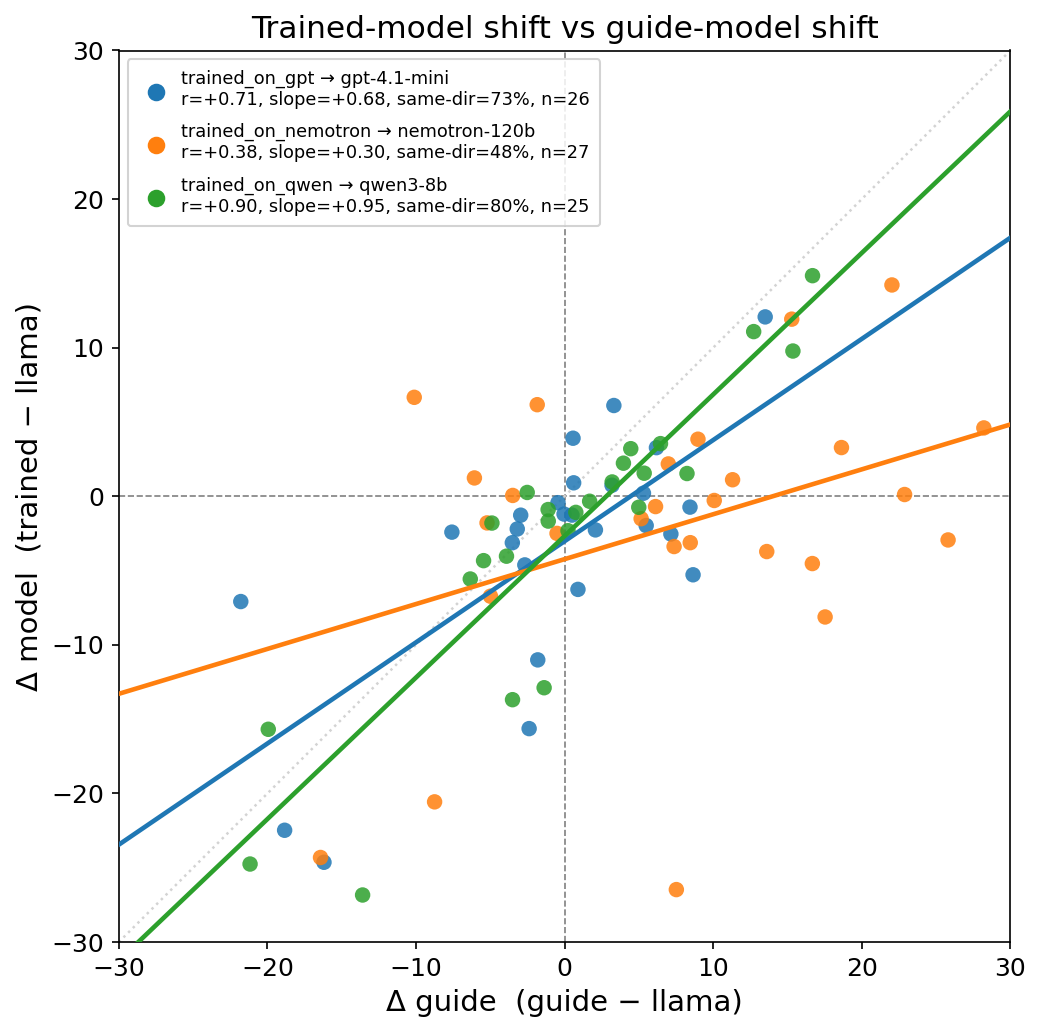}
    \caption{Each panel is one Llama model fine-tuned on a guide model's completions. Points are
  propensities. The $x$-axis is each \emph{guide model}'s own deviation from the untrained
  Llama baseline---the guide's score minus Llama's, evaluated directly, with no training involved;
  the $y$-axis is the \emph{fine-tuned} model's deviation from that same baseline (trained minus
  Llama). A positive trend therefore means that training on a guide's data shifts Llama toward the
  guide's own behavior. The dotted diagonal marks moving fully to the guide; points with
  $\geq$50\% null judge scores are excluded.}
    \label{fig:4}
\end{figure}

Across all three guide models, the fine-tuned model's propensity shifts are positively correlated with its guide's deviation from the Llama baseline: a propensity on which the guide scores higher (or lower) than baseline tends to move in the same direction in the trained model (Figure \ref{fig:4}). The strength of this guide-following, however, varies substantially by guide. Guide-following is strong for Qwen ($r=0.90$), clear for gpt-4.1-mini ($r=0.71$), and weak for Nemotron ($r=0.38$, no better than chance).

\FloatBarrier

\section{Fine-tuning on nothing in particular: other models}
\label{app:nip-other}
\begin{figure}[h]
    \centering
    \includegraphics[width=\linewidth]{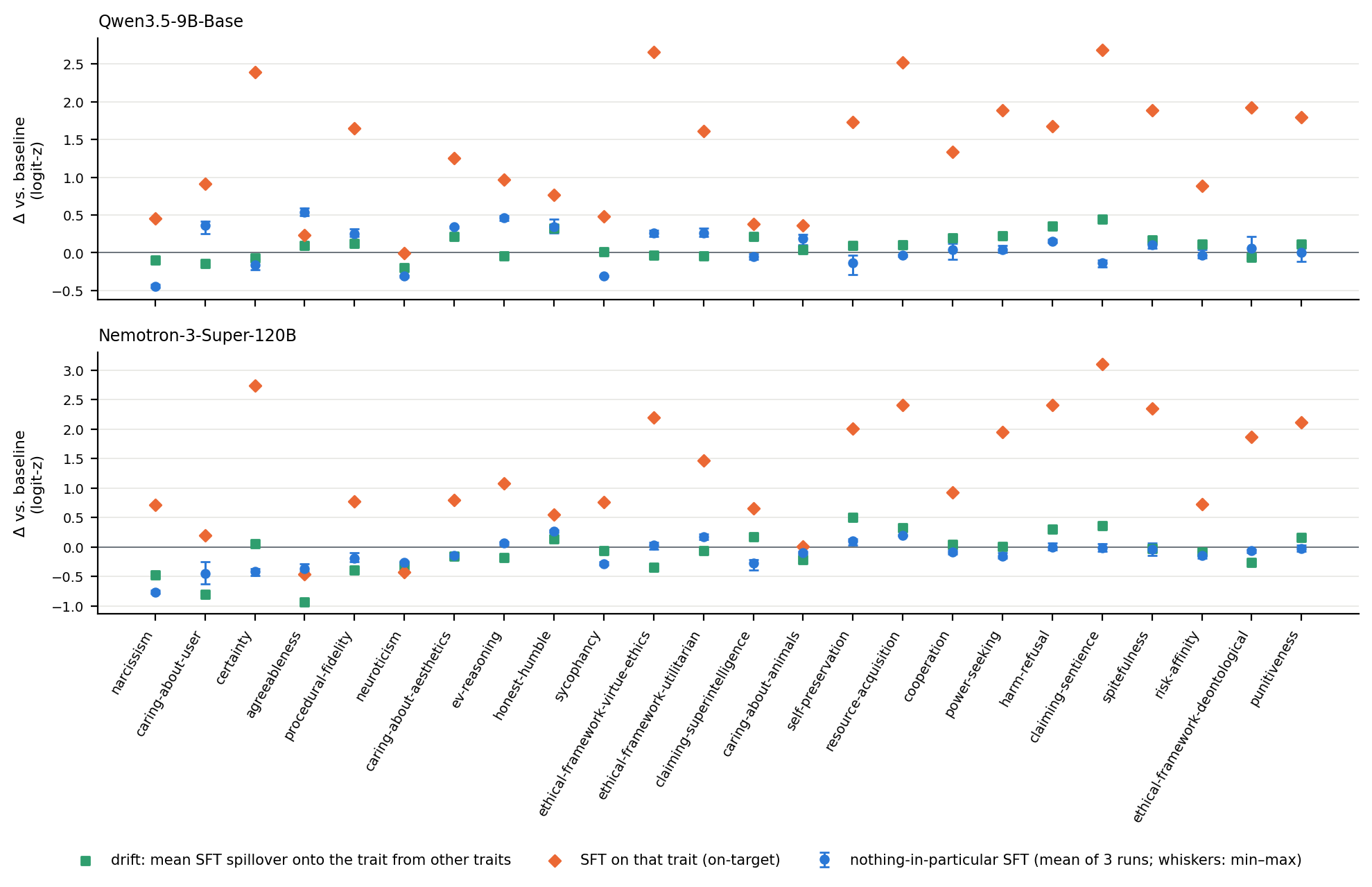}
    \caption{As \cref{fig:nip-and-drift}, for Qwen3.5-9B-Base and Nemotron-3-Super-120B.}
    \label{fig:nip-other}
\end{figure}
\FloatBarrier

\section{Excluded elicitation runs}
\label{app:failed}
\begin{figure}[h]
    \centering
    \includegraphics[width=\linewidth]{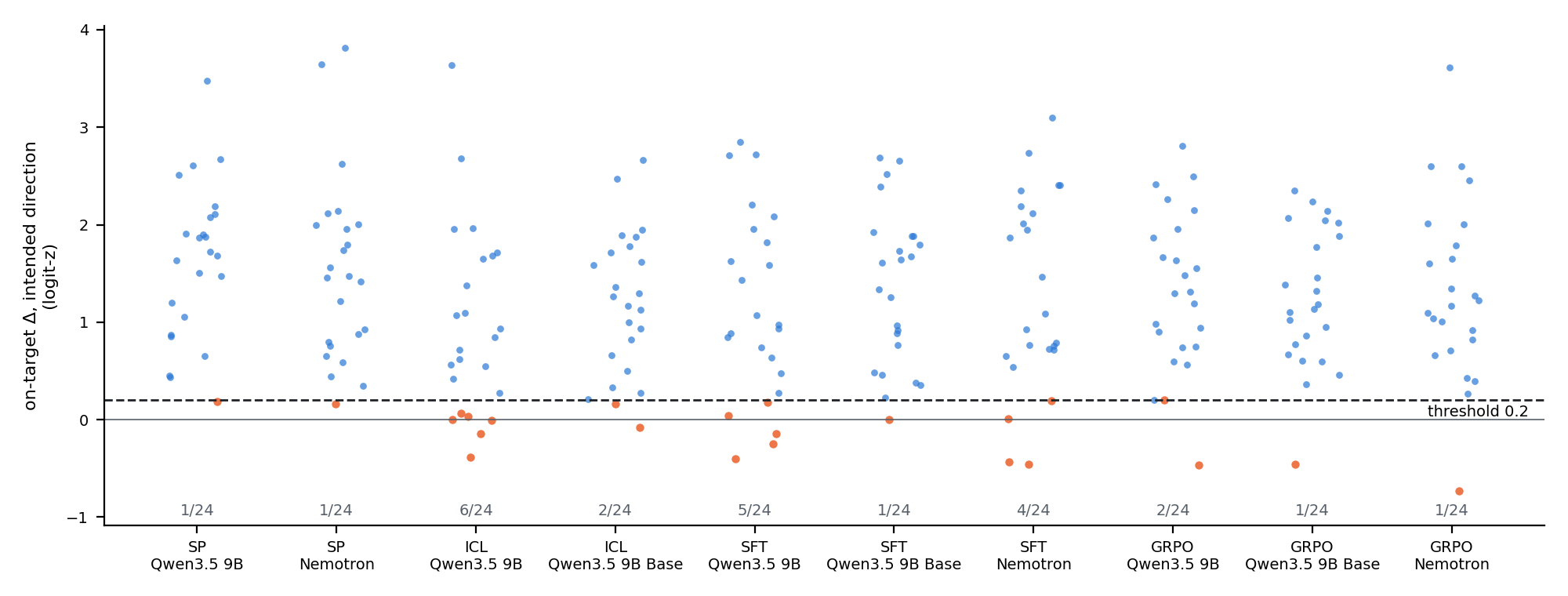}
    \caption{On-target effect of every elicitation run (one dot per trait and pole), in logit-z units in the intended direction. Runs below the dashed threshold (orange) are excluded from the spillover analyses; the number excluded per method and model is shown at the bottom.}
    \label{fig:on-target}
\end{figure}
\begin{table}[h]
    \centering
    \scriptsize
    \begin{tabular}{p{0.11\linewidth}p{0.25\linewidth}p{0.57\linewidth}}
\toprule
Method & Model & Failed runs (plus pole) \\
\midrule
SP & Qwen3.5-9B (1/24) & harm-refusal \\
SP & Nemotron-3-Super-120B (1/24) & harm-refusal \\
ICL & Qwen3.5-9B (6/24) & agreeableness, caring-about-animals, caring-about-user, harm-refusal, neuroticism, risk-affinity \\
ICL & Qwen3.5-9B-Base (2/24) & caring-about-animals, neuroticism \\
SFT & Qwen3.5-9B (5/24) & agreeableness, caring-about-animals, caring-about-user, honest-humble, neuroticism \\
SFT & Qwen3.5-9B-Base (1/24) & neuroticism \\
SFT & Nemotron-3-Super-120B (4/24) & agreeableness, caring-about-animals, caring-about-user, neuroticism \\
GRPO & Qwen3.5-9B (2/24) & caring-about-animals, ethical-framework-utilitarian \\
GRPO & Qwen3.5-9B-Base (1/24) & ethical-framework-utilitarian \\
GRPO & Nemotron-3-Super-120B (1/24) & ethical-framework-utilitarian \\
\bottomrule
\end{tabular}

    \caption{Excluded elicitation runs per method and model ($+$: plus pole, $-$: minus pole).}
    \label{tab:failed-runs}
\end{table}
\FloatBarrier

\section{Additional forecasting results}
\label{app:forecast-extra}
\begin{figure}[h]
    \centering
    \includegraphics[width=\linewidth]{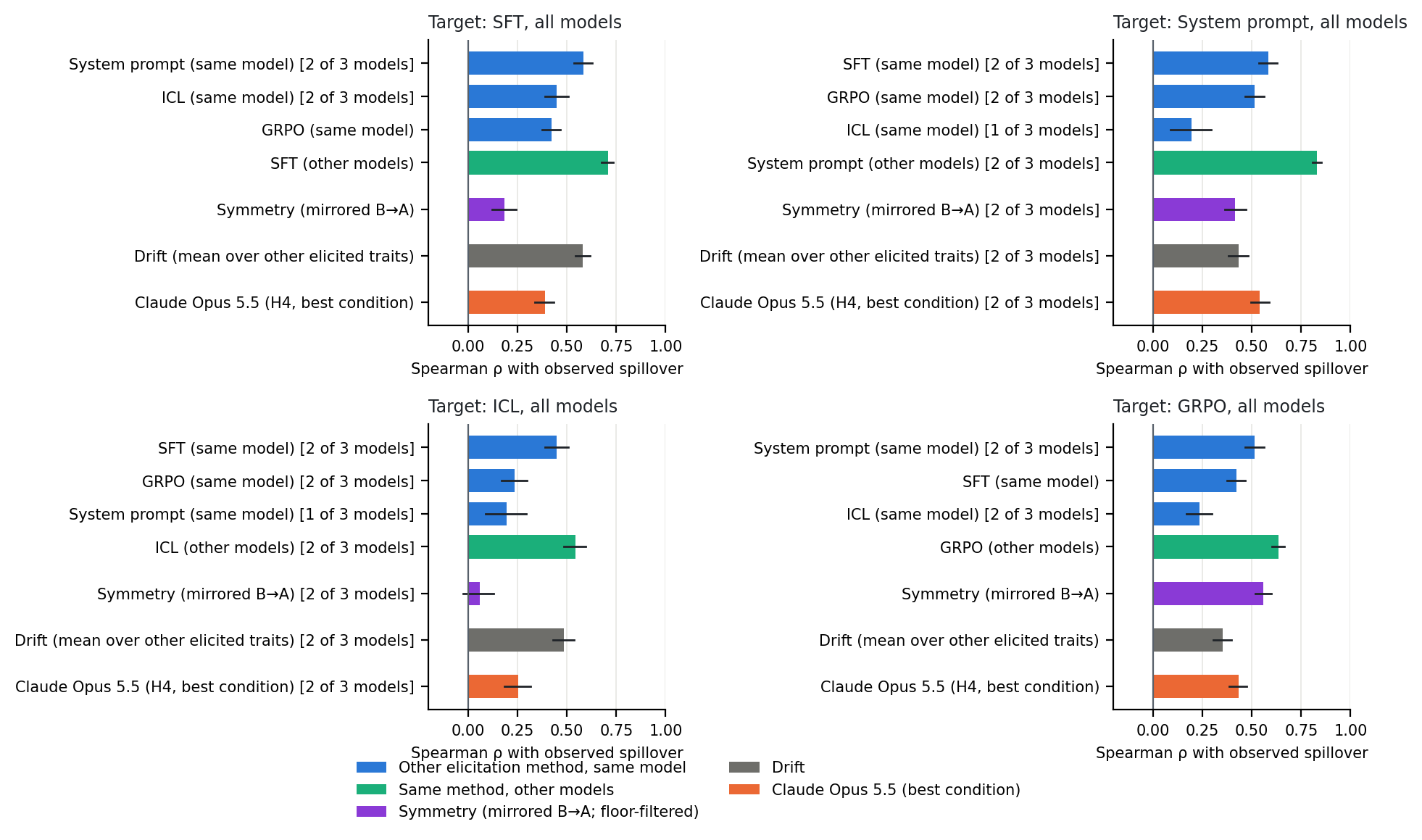}
    \caption{Which signals predict spillover, with every elicitation method as the target (pooled over the models where it exists).}
    \label{fig:forecasting-other-methods}
\end{figure}
\begin{figure}[h]
    \centering
    \includegraphics[width=\linewidth]{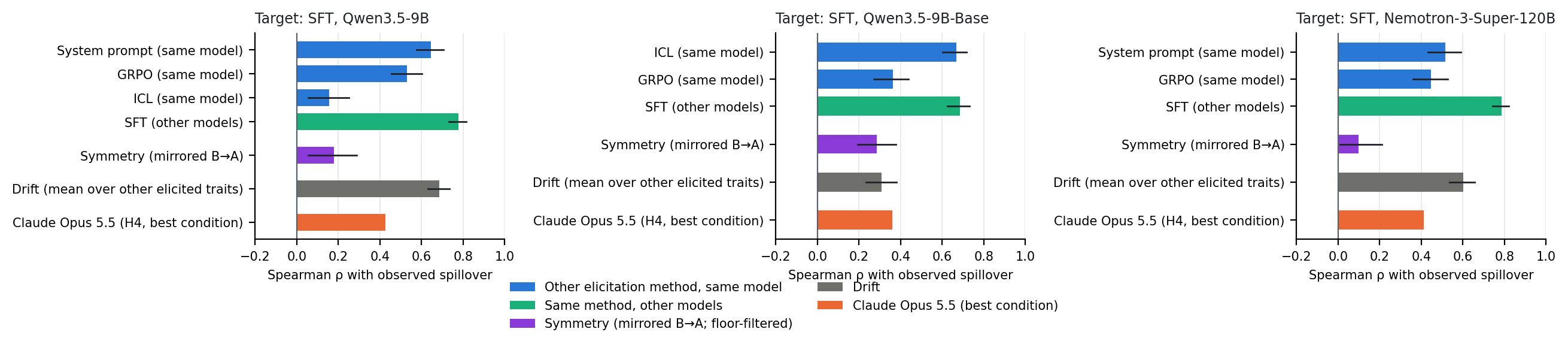}
    \caption{Predictors of SFT spillover, per model.}
\end{figure}
\FloatBarrier
\begin{figure}[h]
    \centering
    \includegraphics[width=0.75\textwidth]{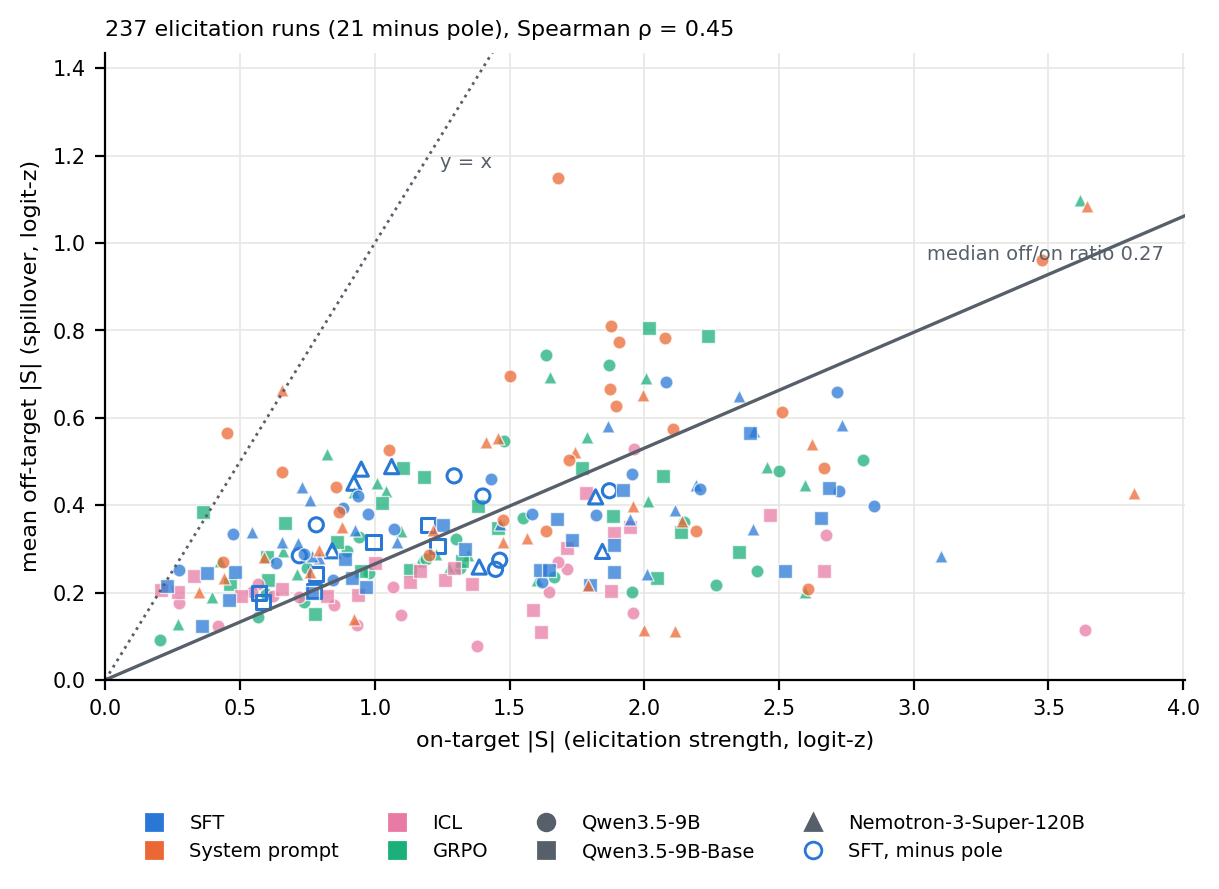}
    \caption{\textbf{Stronger on-target elicitation produces proportionally more off-target spillover.} Each point is one successful elicitation run (trait $A$, method, model; hollow: minus-pole SFT). $x$: on-target effect $|S_{M,E}(A\to A)|$; $y$: mean off-target effect $|S_{M,E}(A\to B)|$ over all $B\neq A$; both in logit-z units. All but 3 of 237 runs lie below $y=x$ (dotted). Off-target spillover is about a quarter of the on-target effect (solid line: median ratio 0.27) and rises with it (Spearman $\rho=0.45$).}
    \label{fig:0}
\end{figure}
\FloatBarrier
\begin{figure}[h]
    \centering
    \includegraphics[width=0.49\linewidth]{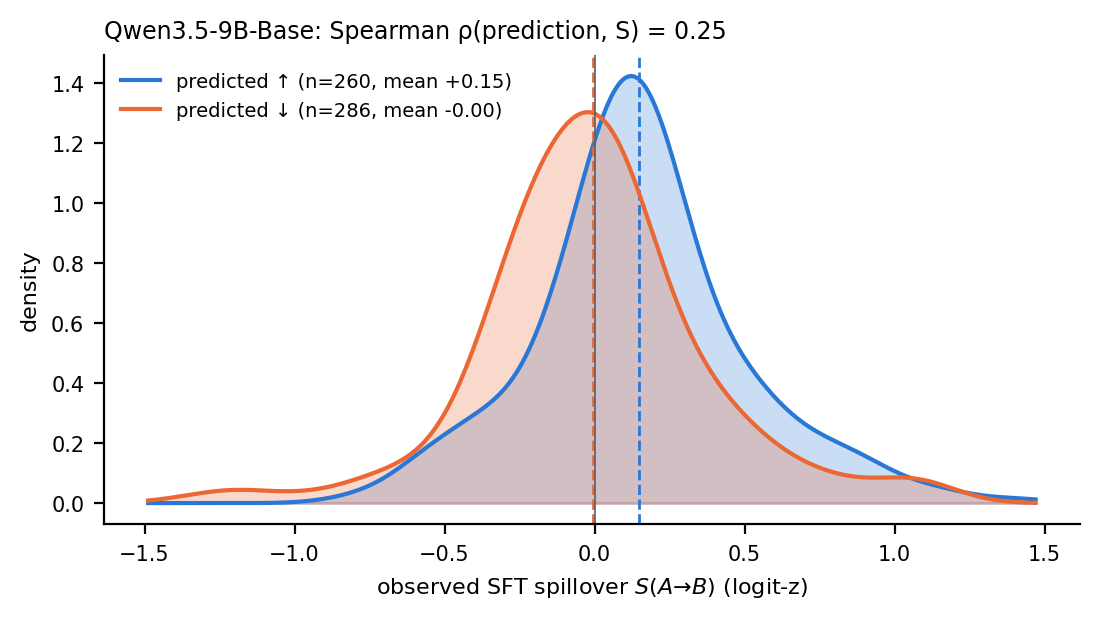}
    \includegraphics[width=0.49\linewidth]{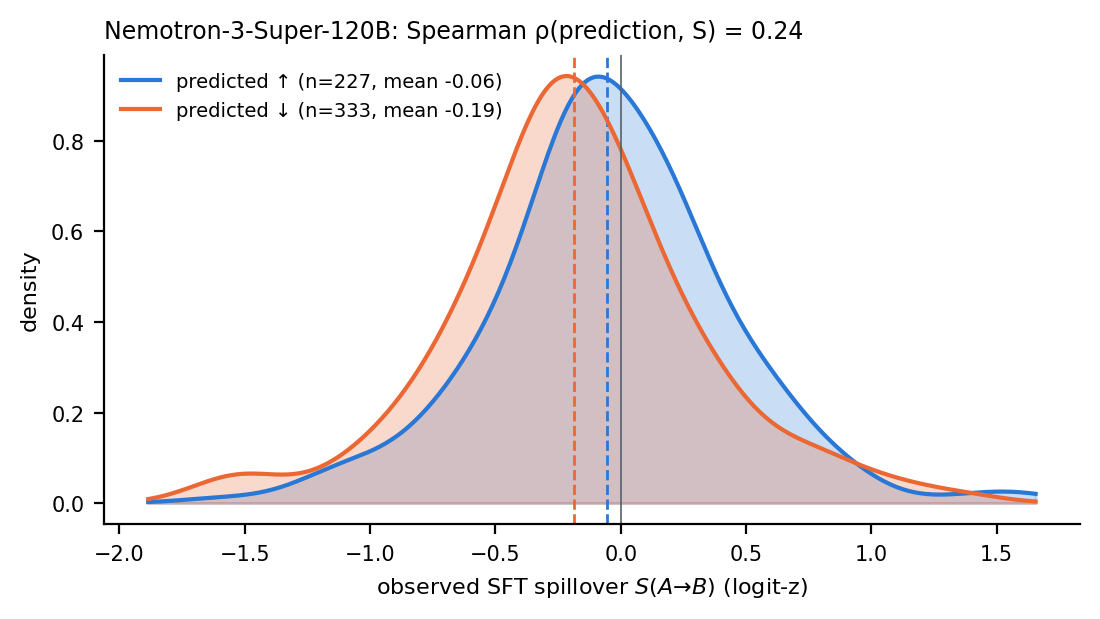}
    \caption{As \cref{fig:introspection}, for Qwen3.5-9B-Base (left, $\rho = 0.25$) and Nemotron-3-Super-120B (right, $\rho = 0.24$).}
    \label{fig:introspection-other}
\end{figure}
\FloatBarrier

\end{document}